\documentclass{article}

\usepackage[preprint]{neurips_2026}

\usepackage[utf8]{inputenc}
\usepackage[T1]{fontenc}
\usepackage{hyperref}
\usepackage{url}
\usepackage{booktabs}
\usepackage{amsfonts}
\usepackage{amsmath}
\usepackage{amssymb}
\usepackage{amsthm}
\usepackage{nicefrac}
\usepackage{microtype}
\usepackage{xcolor}
\usepackage{graphicx}
\graphicspath{{figures/}}
\usepackage{multirow}
\usepackage{bm}
\usepackage{enumitem}

\newtheorem{proposition}{Proposition}

\title{Dendritic In-Context Learning in a Single-Layer Spiking Neural Network}

\author{
Juwei Shen \\
  Department of Computing, FCMS\\
  The Hong Kong Polytechnic University\\
  Hong Kong, China \\
  {\tt\small sheldon.shen@connect.polyu.hk}
\And
Yujie Wu\\
  Department of Computing, FCMS\\
  The Hong Kong Polytechnic University\\
  Hong Kong, China \\
  {\tt\small yu-jie.wu@polyu.edu.hk} \\
\And
Changwen Chen\thanks{Corresponding Author}\\
  Department of Computing, FCMS\\
  The Hong Kong Polytechnic University\\
  Hong Kong, China \\
  {\tt\small changwen.chen@polyu.edu.hk} \\
}

\begin{document}

\maketitle

\begin{abstract}In-context learning (ICL) operates via implicit gradient descent embedded in the forward pass of modern AI architectures --- Transformers~\citep{akyurek2023icl,vonoswald2023transformers}, Mamba~\citep{park2024mamba}, state-space models~\citep{sushma2024ssm}, and MLPs~\citep{tong2025mlp}. Capturing this capability in biologically plausible Spiking Neural Networks (SNNs) has remained an open challenge: existing SNNs fail the Garg-2022 benchmark at non-trivial task dimensions. We trace this failure to a \emph{structural assumption}: prior SNN designs route adaptation through inference-time synaptic plasticity, viewing the dendritic compartment as a passive conduit for error or teacher signals. \textbf{We challenge this assumption.} The subthreshold dynamics of a single dendritic compartment already implement a complete online learning algorithm. By treating the compartment as the computational substrate rather than a passive conduit, we propose \textbf{DendriCL} --- a single-layer compartmental spiking architecture whose apical recurrence is structurally identical to leaky online Widrow--Hoff LMS~\citep{widrow1960}. This dynamics-only update collapses the architectural depth required for general-purpose ICL to a single layer. DendriCL is uniquely seed-stable at super-dimensional Garg-2022 ICL --- where dense Transformers exhibit grokking-style instability and fail past moderate task dimension --- and a linear probe recovers the reference online-LMS trajectory directly from the apical membrane at $R^2 = 0.93$, showing the algorithm is structurally embedded in the dynamics rather than implicitly discovered during training. Taken together, ICL requires neither attention, depth, nor inference-time plasticity: a single compartment with online-LMS dynamics is sufficient.
\end{abstract}

\begin{figure}[t]
\centering
\includegraphics[width=\linewidth]{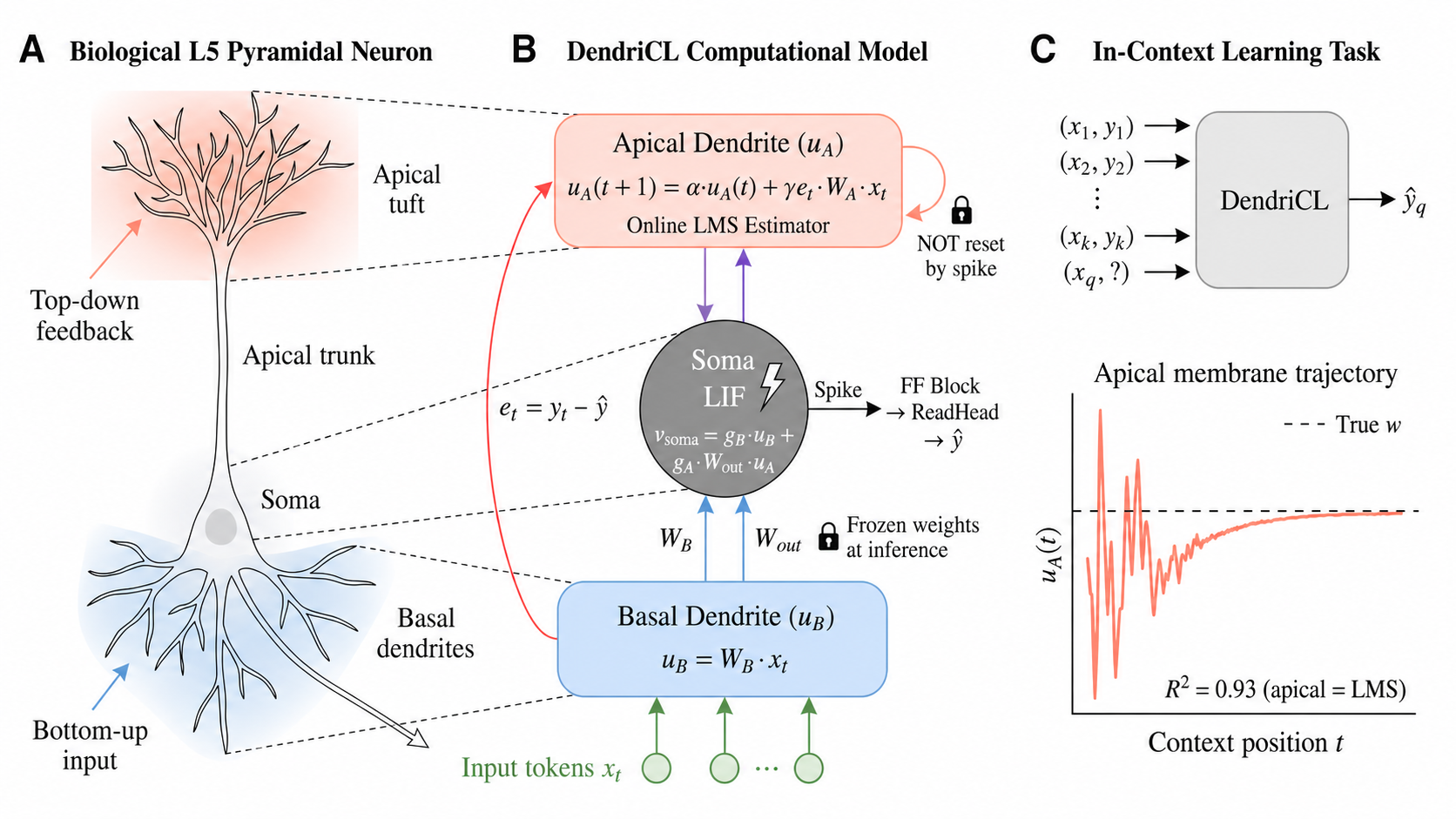}
\caption{\textbf{DendriCL overview.} \emph{(A)} Biological layer-5 cortical pyramidal neuron: apical tuft receives top-down feedback, basal dendrites receive bottom-up input, and the soma integrates both. \emph{(B)} DendriCL computational model --- a single compartmental layer implementing the apical-basal-soma architecture with a learned online-LMS update in the apical membrane potential $\bm{u}_A$. Synaptic weights $W_A, W_B, W_{\text{out}}$ are \emph{frozen at inference}; the apical state is not reset by spikes and evolves across the full context. \emph{(C)} Garg~2022 ICL task: given $k$ labeled pairs $(\bm{x}_i, y_i)$ the model predicts $\hat{y}_q$ for a query $\bm{x}_q$. The apical membrane trajectory converges to the task parameter $\bm{w}$ with linear-probe decoding $R^2 = 0.93$ into the reference online-LMS estimate.}
\label{fig:overview}
\end{figure}

\section{Introduction}
\label{sec:intro}

\textbf{In-context learning} (ICL) --- the ability of a trained sequence model to solve a new task from a handful of labeled examples in its prompt, without any parameter update~\citep{brown2020language} --- has been mechanistically traced, across architecture families, to an \emph{implicit gradient-descent algorithm} embedded in the trained forward pass. On the synthetic function-class protocol of \citet{garg2022}, \citet{akyurek2023icl} showed that trained Transformers internally recover the iterates of a gradient-based learner; \citet{vonoswald2023transformers} made this equivalence explicit (linear self-attention $\equiv$ one GD step on linear regression; stacking $\equiv$ trajectory); \citet{ahn2024transformers} sharpened it to \emph{preconditioned} GD. The reading has since generalized well beyond attention: state-space models~\citep{sushma2024ssm}, Mamba~\citep{park2024mamba,grazzi2024mamba}, MLPs and MLP-Mixers~\citep{tong2025mlp}, and linear Transformers~\citep{schlag2021linear,vladymyrov2024linear} all exhibit the same implicit-GD signature. ICL is now understood as a \emph{generic} property of trained sequence models, not an attention-specific artifact.

One substrate has remained conspicuously outside this picture: \textbf{spiking neural networks} (SNNs), the substrate closest to biological computation and the native target of neuromorphic hardware~\citep{davies2018loihi}. To our knowledge, no prior work has demonstrated general-purpose ICL in an SNN on the Garg protocol; our own survey of canonical SNN architectures confirms a wide gap (Spikformer~\citep{zhou2023spikformer}: $R^2 = 0.72$ at $d{=}20$; Pure LIF: $R^2 \approx 0.09$; LSNN~\citep{bellec2018lsnn}: $R^2 \approx 0.01$; Spiking SSMs at the chance floor). \emph{Is this incidental --- a matter of insufficient parameters, training compute, or architectural search --- or structural?}

\textbf{We argue it is structural, and that the structure responsible has been misidentified.} Implicit-GD requires, somewhere in the forward pass, a state vector with dimensionality comparable to the task dimension that is updated by a gradient-like rule. A standard leaky-integrate-and-fire neuron carries a \emph{scalar} membrane potential, reset after each spike --- no persistent, multi-dimensional subthreshold state, no candidate substrate for the algorithm. Existing biology-inspired approaches sidestep this by routing adaptation through inference-time synaptic plasticity~\citep{bellec2018lsnn,miconi2018,shen2025} and treating the dendritic compartment as a passive carrier of driving signals: apical as teacher~\citep{urbanczik2014}, as backpropagated error~\citep{sacramento2018}, as attentional gate~\citep{larkum2013}, as external task-ID input~\citep{iyer2022}. \emph{This routing is unnecessary.} The subthreshold dynamics of a single dendritic compartment alone implement a complete online learning algorithm: with all synaptic weights frozen at inference, the compartment is the substrate of the algorithm, not a conduit for it.

The architectural ingredient is anatomically standard. From the perspective of \emph{single-neuron computation}, a cortical layer-5 pyramidal neuron carries, in its \emph{apical dendrite} (the upper input branch integrating top-down signals), a persistent multi-dimensional subthreshold voltage with calcium plateaus on 100+\,ms timescales~\citep{larkum1999,larkum2002,gidon2020,larkum2013,major2013} --- precisely the missing substrate. We instantiate this in \textbf{DendriCL}: a \emph{single}-layer compartmental spiking network whose apical recurrence
\[
\bm{u}_A(t+1) = \alpha\, \bm{u}_A(t) + \gamma\, (y_t - \hat{y}_t)\, W_A\, \bm{x}_t
\]
is structurally identical to leaky online Widrow--Hoff LMS~\citep{widrow1960}. Once $(\alpha, \gamma, W_A, W_B)$ are trained end-to-end by BPTT, the apical membrane provably tracks the task parameter $\bm{w}$, and every synaptic weight is frozen at inference.

\textbf{Contributions.} We position the apical compartment as an \textbf{active online estimator with frozen synapses}, in contrast to the plasticity-driven interpretations of prior compartmental models~\citep{urbanczik2014,sacramento2018,larkum2013,iyer2022}. On this basis we introduce \textbf{DendriCL}, the first \emph{single}-layer compartmental SNN to solve general-purpose Garg-2022 ICL across $d \in \{5, \ldots, 50\}$ where canonical SNN baselines (Spikformer, Pure LIF, LSNN, Spike-driven~V2) collapse at $d \geq 20$, and the only architecture that is seed-stable at $d \geq 30$ where dense Transformers exhibit grokking-style bimodality and fail past $d{=}40$. A linear probe recovers the reference online-LMS trajectory directly from the apical membrane at $\bm{R^2 = 0.93}$, giving direct mechanistic verification of the apical-LMS equivalence. The single-layer compartmental design yields a $\sim\!4\times$ spike reduction over Pure LIF and a projected $\sim\!10\times$ Loihi-class energy advantage; to our knowledge, this is the first ICL setting in which architectural simplicity and inference-time efficiency co-vary rather than trade off.

\section{Background and Related Work}
\label{sec:related}

\paragraph{ICL mechanisms across substrates.} The mechanistic study of ICL begins with \citet{garg2022}, who isolated the phenomenon in a synthetic function-class setting. On this benchmark, \citet{akyurek2023icl} used linear probes to recover iterates of least-squares and gradient-descent solvers from trained-Transformer hidden states. \citet{vonoswald2023transformers} proved that a linear self-attention layer can implement one step of GD on a regression loss, with stacked layers yielding the trajectory; \citet{ahn2024transformers} extended this to preconditioned GD. \citet{schlag2021linear} earlier showed that linear Transformers are equivalent to fast-weight programmers, a connection generalized by \citet{vladymyrov2024linear}. Beyond attention, the implicit-GD reading has been demonstrated for state-space models~\citep{sushma2024ssm}, Mamba~\citep{park2024mamba,grazzi2024mamba}, and MLPs/MLP-Mixers~\citep{tong2025mlp}. Our work positions a single compartmental spiking layer as the spiking-substrate analog of this lineage.

\paragraph{Biologically-plausible learning and compartmental neurons.} A largely non-overlapping line studies how biological circuits could plausibly implement learning algorithms. \citet{urbanczik2014} cast the apical compartment as a \emph{teacher signal} driving synaptic plasticity toward the somatic firing rate. \citet{sacramento2018} extended this to a three-compartment microcircuit in which the apical receives a \emph{backpropagated error} from local interneurons. \citet{miconi2018} introduced differentiable plasticity --- Hebbian fast weights adapting at inference. \citet{bellec2018lsnn} added a scalar adaptive threshold to LIF (LSNN) for short-term memory; \citet{iyer2022} proposed Active Dendrites with externally-provided task-context gates; \citet{shen2025} (IP$^2$-RSNN) combined intrinsic time-constant plasticity with second-order meta-learning. A structural commonality unites these proposals: adaptation is driven by \emph{parameter changes} at inference time, never by the subthreshold dynamics of a frozen-weight network. None has been evaluated on Garg-style ICL.

\paragraph{Dynamics-based test-time adaptation.} Closer in spirit are architectures that adapt via continuous-time dynamics with frozen weights. The \emph{Liquid State Machine}~\citep{maass2002lsm} projects inputs through a random recurrent spiking reservoir and trains only a linear readout; DendriCL retains the frozen-weight philosophy but replaces the random reservoir with a structured apical compartment whose dynamics admit an explicit algorithmic interpretation (linear probe $R^2 = 0.93$). Echo State Networks~\citep{jaeger2001esn} and FORCE learning~\citep{sussillo2009force} share the trained-readout/frozen-recurrence design but were not evaluated on Garg-style ICL; Liquid Time-Constant~\citep{hasani2020ltc} and CfC networks~\citep{hasani2022cfc} use input-gated time constants for sequential modeling but are neither spiking nor compartmental. Den\`{e}ve's spiking Bayesian filters~\citep{deneve2008} address scalar estimation rather than multi-dimensional ICL.

\paragraph{Scope: SNN methods excluded by the frozen-weight criterion.} A principled ICL comparison requires every parameter to remain frozen at test time, so any adaptation visible in the forward pass is mechanistic rather than ordinary training. This excludes inference-time-plasticity SNNs --- e-prop~\citep{bellec2020eprop}, differentiable plasticity~\citep{miconi2018}, IP$^2$-RSNN~\citep{shen2025}, STDP-based methods, and hardware-local-rule learning on Loihi~\citep{davies2018loihi} and SpiNNaker --- which solve the related but distinct problem of \emph{how} to update parameters cheaply at inference. We view them as complementary to DendriCL.

\section{Method}
\label{sec:method}

\subsection{Architecture}

DendriCL is a single compartmental spiking layer of $d_{\text{model}}=384$ parallel pyramidal-like units. Each unit has a \emph{basal} dendrite (feedforward projection of the input), an \emph{apical} dendrite carrying a persistent multi-dimensional subthreshold voltage $\bm{u}_A \in \mathbb{R}^{d_{\text{apical}}}$, and an \emph{LIF soma} that integrates both compartments and emits spikes. Inputs are $k+1$ tokens, each encoding $[\bm{x}_i; y_i; \text{flag}_i]$ with $\text{flag}_i \in \{0,1\}$ marking the query position. The full per-unit recurrence at context position $t$ is:
\begin{align}
\bm{u}_B(t) &= W_B\, \bm{x}_t & &\text{(basal projection)} \nonumber \\
\hat{y}_t &= \bm{u}_A(t)^\top (W_A \bm{x}_t) & &\text{(scalar prediction)} \nonumber \\
e_t &= (1 - \text{flag}_t)(y_t - \hat{y}_t) & &\text{(gated error)} \nonumber \\
\bm{u}_A(t+1) &= \alpha\, \bm{u}_A(t) + \gamma\, e_t\, W_A\, \bm{x}_t & &\text{(apical online-LMS)} \\
v_{\text{soma}}(t) &= g_B\, \bm{u}_B(t) + g_A\, W_{A,\text{out}}\, \bm{u}_A(t) & &\text{(somatic integration)} \nonumber \\
s(t) &= \mathbb{1}[v_{\text{soma}}(t) > \theta],\ v_{\text{soma}} \leftarrow v_{\text{soma}} - \theta\, s(t) & &\text{(LIF spike, soft reset)} \nonumber
\end{align}
The apical state $\bm{u}_A$ is initialized to zero and \emph{not reset} by the somatic spike; it evolves continuously across the full context. The error $e_t$ is gated off at the query position. A small post-LIF block ($\text{FF}_1 \to \text{LIF} \to \text{FF}_2$) and a linear readout active only at the query position produce $\hat{y}_q$. All trainable parameters --- $\alpha, \gamma, W_A, W_B, W_{A,\text{out}}, g_A, g_B, \theta$ plus FF weights --- are learned end-to-end by BPTT; no parameter is updated at inference. DendriCL has $d_{\text{model}}=d_{\text{apical}}=384$ and $\approx\!0.75$M total parameters.

\begin{figure}[h]
\centering
\includegraphics[width=\linewidth]{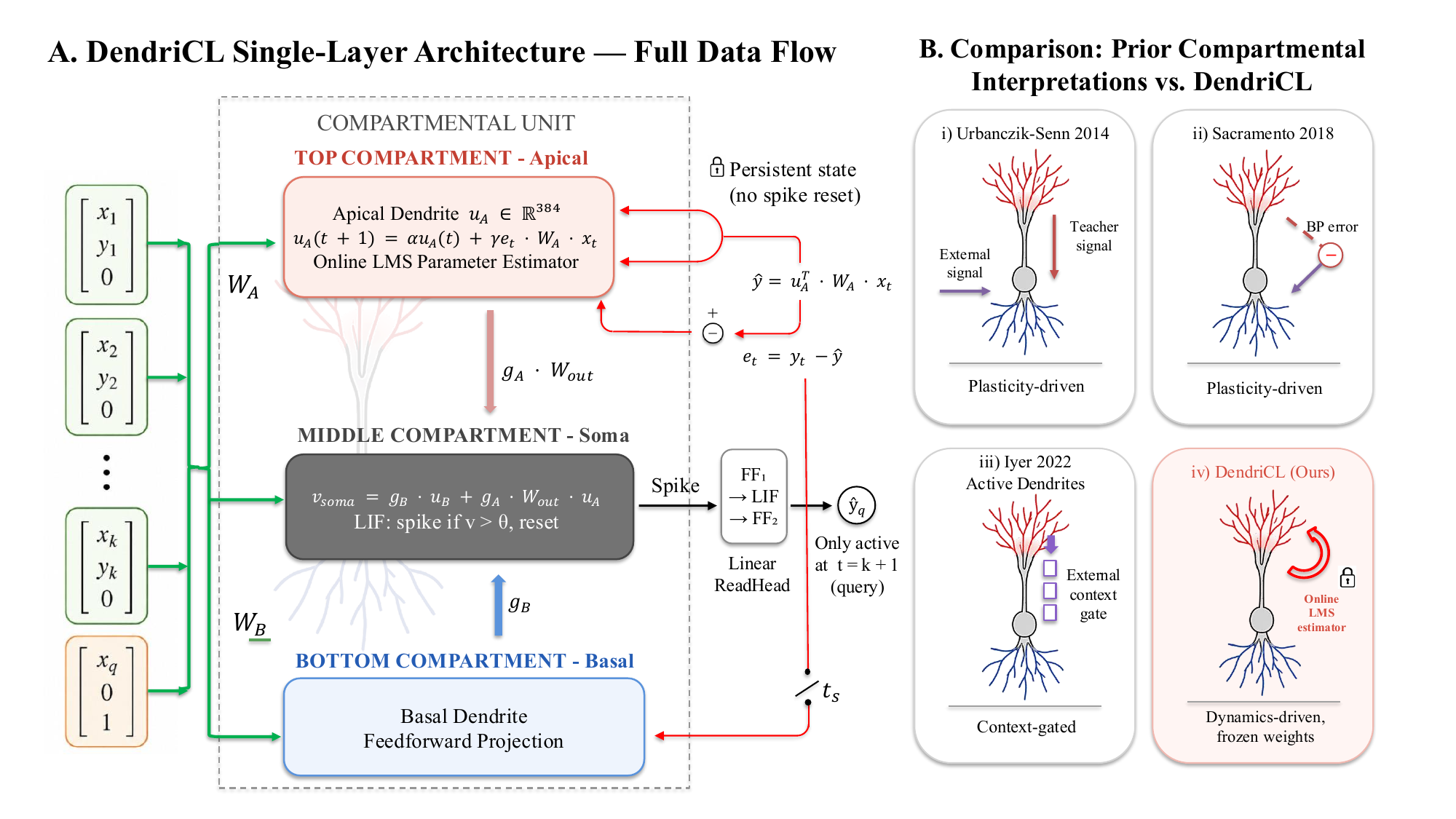}
 \caption{\textbf{DendriCL architecture and contrast with prior compartmental models.} \emph{(A)} Single-layer data flow: input tokens $[\bm{x}_i; y_i; \text{flag}_i]$ feed both the basal projection $\bm{u}_B = W_B \bm{x}_t$ and the apical online-LMS recurrence. The LIF soma integrates $v_{\text{soma}} = g_B \bm{u}_B + g_A W_{\text{out}} \bm{u}_A$ and spikes at threshold; a readout is active only at the query position. \emph{(B)} Contrast with four prior compartmental-neuron interpretations: Urbanczik-Senn~2014 (apical = teacher signal, plasticity-driven), Sacramento~2018 (apical = backprop error, plasticity-driven), Iyer~2022 Active Dendrites (apical segments = external context gate), DendriCL (apical = online LMS estimator, \emph{dynamics-driven with frozen weights}).}
\label{fig:arch}
\end{figure}

\subsection{Training}

All parameters --- $\alpha, \gamma, W_A, W_B, W_{A,\text{out}}, g_A, g_B, \theta$, plus FF weights --- are trained from scratch via backpropagation through time (BPTT). Each minibatch samples a fresh task parameter $\bm{w} \sim \mathcal{N}(0, I_d)$, a fresh context, and a fresh query. Loss is computed only at the query position. LIF surrogate gradients use the arctan approximation~\citep{neftci2019surrogate}. Optimizer: AdamW, learning rate $10^{-3}$, weight decay $10^{-4}$, cosine schedule, batch size 64, 10k steps (baseline) or 50k steps (compute-matched); 3 seeds per configuration (5 seeds for $d \in \{20, 30\}$). \emph{No pre-training, no plasticity at inference.}

\subsection{Core Mechanism: Apical $\equiv$ Leaky Online LMS}

Setting $W_A = I$ and absorbing the output projection into the readout, the apical update reduces to classical leaky Widrow-Hoff LMS:
\[
\hat{\bm{w}}_{t+1} = \alpha \hat{\bm{w}}_t + \gamma (y_t - \hat{\bm{w}}_t^\top \bm{x}_t) \bm{x}_t.
\]
\begin{proposition}[Informal; proof in Appendix~\ref{app:proof}]
\label{prop:lms}
Under $\bm{x}_t \sim \mathcal{N}(0, I_d)$ i.i.d.\ and $y_t = \bm{w}^\top \bm{x}_t + \epsilon_t$ with $\epsilon_t \sim \mathcal{N}(0, \sigma^2)$, for $0 < \gamma < 2/d$ and $\alpha$ suitably scheduled, the apical state satisfies $\mathbb{E}\|\bm{u}_A(k) - \bm{w}\|^2 = \mathcal{O}(d/k)$.
\end{proposition}
This is the classical LMS convergence theorem~\citep{widrow1960,sayed2003adaptive}. Our contribution is structural: the update is embedded in the compartmental architecture, and BPTT tunes $(\alpha, \gamma, W_A)$ to make the built-in rule optimal across the task distribution. The linear-probe analysis (\S\ref{sec:probe}) empirically confirms the correspondence. Our result is a spiking-substrate analog of the dynamics-as-algorithm theorems of~\citep{vonoswald2023transformers,sushma2024ssm,schlag2021linear}. \textbf{Rigor caveat}: the analysis covers the apical compartment (purely linear); the LIF reset introduces a nonlinearity our theorem does not formally handle. Empirical probe analysis ($R^2=0.93$) establishes that the correspondence holds despite this nonlinearity.

\paragraph{Biological grounding and distinction from prior compartmental models.}
DendriCL's three-compartment apical-basal-soma layout is anatomically grounded in cortical layer-5 pyramidal neurons~\citep{larkum2013,major2013,larkum1999,gidon2020}; the assignment of the apical compartment to error-driven dynamics is consistent with predictive coding~\citep{rao1999,bastos2012}; and the specific online-LMS mechanism is a computational hypothesis falsifiable by future \emph{in vivo} recording. DendriCL is also structurally distinct from six prior compartmental models~\citep{urbanczik2014,sacramento2018,miconi2018,bellec2018lsnn,iyer2022,shen2025} along the dimensions of adaptation substrate and inference-time plasticity. We defer the three-tier biological evidence breakdown and the full compartmental-model comparison table (Table~\ref{tab:compartmental-compare}) to Appendix~\ref{app:bio-distinction}.

\section{Experimental Setup}
\label{sec:exp}

\textbf{Tasks.} Garg~2022 linear regression at $d \in \{5, 10, 15, 20, 25, 30, 40, 50\}$ with $k = 2d$; binary classification ($d=10, k=20$); 2-layer ReLU NN regression ($d=20, k=40$).

\textbf{Architectures.} We evaluate 15 architectures spanning attention-based (Transformer, Spikformer, Spike-driven~V2, Linear Transformer), recurrent (GRU, LSNN, LTC), state-space (SpikingSSM), spiking (Pure LIF), MLP-family (Pure MLP, MLP-Mixer), and compartmental (Active Dendrites, DendriCNN, DendriCL with $L \in \{1, 2\}$) at matched 0.6--1.3M parameter budgets.

\textbf{Training budget.} We report at two regimes: \emph{10k steps} (baseline) and \emph{50k steps} (5$\times$ compute-matched). The 50k regime probes whether failures at high $d$ are training-budget artifacts or architectural limits. Per-architecture hyperparameters (optimizer, learning rate, depth, width) and hardware/wall-clock details are in Appendices~\ref{app:arch-hparams} and~\ref{app:additional-details}.

\textbf{Evaluation.} Best eval $R^2$ on 1500 held-out tasks per checkpoint, averaged across 3 seeds (5 seeds at $d \in \{20, 30\}$, 6 seeds for Transformer at $d=30$ to characterize bimodality).

\section{Results}
\label{sec:results}

\subsection{Main Comparison at \texorpdfstring{$d=10, d=20$}{d=10, d=20}}

Table~\ref{tab:main} compares all baseline architectures at two task dimensionalities. At $d=10$ all but LSNN cluster near $R^2 \approx 0.8$, indicating that ICL itself is broadly attainable across architectures at low dimensionality. At $d=20$ the picture changes sharply: a clean bifurcation separates four \emph{ICL-capable} architectures (Transformer, Spikformer, DendriCL, DendriCNN) at $R^2 \geq 0.72$ from four that collapse to near chance ($R^2 \leq 0.09$: Pure LIF, Active Dendrites, GRU, LSNN). DendriCL is the highest-scoring spike-based model and is within $3$\,pp of its non-spiking ablation DendriCNN, indicating spike binarization itself is not the limiting factor. The full 15-architecture $\times$ 8-dimension $\times$ 2-compute matrix --- including LSNN, SpikingSSM, LTC, Spike-driven~V2, and the MLP family --- is in Appendix~\ref{app:full-matrix} (Table~\ref{tab:full-matrix}).

\begin{table}[h]
\caption{Main comparison at two task dimensionalities (3 seeds, best eval $R^2$). DendriCL is benchmarked against single-layer biology-inspired SNN architectures (its native tier); Transformer and Spikformer values are shown for cross-tier reference but use dense attention with $\geq 1.3\times$ parameter budget. \textbf{Within tier, DendriCL achieves column-best $R^2$ at both $d=10$ and $d=20$}, while every other single-layer SNN baseline collapses from $R^2 \approx 0.8$ at $d=10$ to near-chance at $d=20$. Bold marks within-tier column maximum; reference rows are not bolded. ($n=3$ seeds, mean $\pm$ std.)}
\label{tab:main}
\centering
\footnotesize
\begin{tabular}{lcccr}
\toprule
Architecture & $d=10$ $R^2$ & $d=20$ $R^2$ & Params \\
\midrule
\multicolumn{4}{l}{\emph{Cross-tier references (dense attention; not directly comparable)}} \\
Transformer (continuous attn) & 0.996 & 0.989 & 1.0M \\
Spikformer (spike attn) & 0.977 & $0.724 \pm 0.059$ & 1.0M \\
\midrule
\multicolumn{4}{l}{\emph{Our method and its continuous-NN ablation}} \\
\textbf{DendriCL (ours, single-layer)} & $\bm{0.807 \pm 0.005}$ & $\bm{0.820 \pm 0.005}^{*}$ & \textbf{0.75M} \\
DendriCL ($L=2$) & $0.811 \pm 0.001$ & $0.793 \pm 0.004$ & 0.75M \\
DendriCNN (continuous ablation) & $0.824 \pm 0.004$ & $0.833 \pm 0.001$ & 0.75M \\
\midrule
\multicolumn{4}{l}{\emph{Single-layer biology-inspired baselines (DendriCL's tier)}} \\
Pure LIF & 0.801 & 0.086 & 1.0M \\
Active Dendrites~\citep{iyer2022} & $0.668 \pm 0.005$ & $0.061 \pm 0.018$ & 1.29M \\
GRU & $0.666 \pm 0.013$ & $0.058 \pm 0.014$ & 0.79M \\
LSNN~\citep{bellec2018lsnn} & $0.011 \pm 0.003$ & $0.008 \pm 0.002$ & 0.62M \\
\bottomrule
\end{tabular}
\\[0.2em]
\raggedright\footnotesize $^{*}$DendriCL $d{=}20$ reported at 50k-step compute (10k: $0.805 \pm 0.009$, $n=5$).
\end{table}

\subsection{DendriCL is uniquely seed-stable at super-$d$}
\label{sec:compute-matched}

To distinguish architectural limits from training-budget artifacts, we retrain all key architectures at $5\times$ compute (50k vs.\ 10k steps). Table~\ref{tab:compute-matched} reports eval $R^2$ at $d \in \{25, 30, 40, 50\}$ for DendriCL, Transformer, Spikformer, Pure LIF, and Spike-driven~V2 across both compute budgets; the corresponding bar comparison and Bayes-optimal ridge ceiling are visualized in Fig.~\ref{fig:compute-matched} (Appendix~\ref{app:figs}).

\begin{table}[h]
\caption{Compute-matched (50k step) comparison at super-$d$. \textbf{DendriCL is the only architecture that maintains $R^2 > 0.5$ across $d \in \{30, 40, 50\}$}; Transformer is column-best at $d=25$ but collapses to chance from $d=30$ onward, exposing an architectural rather than budget-induced ceiling. Bold marks the per-column maximum across all architectures. Numbers are mean $\pm$ std over 3 seeds (6 seeds for Transformer at $d{=}30$).}
\label{tab:compute-matched}
\centering
\footnotesize
\setlength{\tabcolsep}{4pt}
\resizebox{\linewidth}{!}{%
\begin{tabular}{lccccc}
\toprule
Architecture & Compute & $d=25$ & $d=30$ & $d=40$ & $d=50$ \\
\midrule
DendriCL (ours) & 50k & $0.809 \pm 0.010$ & $\bm{0.807 \pm 0.005}$ & $\bm{0.787 \pm 0.005}$ & $\bm{0.649 \pm 0.036}$ \\
DendriCL (ours) & 10k & $0.791 \pm 0.001$ & $0.778 \pm 0.004$ & $0.690 \pm 0.021$ & $0.498 \pm 0.029$ \\
\midrule
Transformer & 50k & $\bm{0.991 \pm 0.002}$ & $0.386 \pm 0.479^\dagger$ & $0.009 \pm 0.001$ & $0.008 \pm 0.001$ \\
Transformer & 10k & $0.58 \pm 0.39$ & $0.023 \pm 0.002$ & $0.008 \pm 0.0001$ & --- \\
\midrule
Spikformer & 50k & $0.656 \pm 0.014$ & $0.636 \pm 0.010$ & $0.501 \pm 0.082$ & $0.301 \pm 0.003$ \\
Spikformer & 10k & --- & $0.424 \pm 0.137$ & $0.064 \pm 0.071$ & $0.011 \pm 0.006$ \\
\midrule
Pure LIF & 50k & --- & $0.394 \pm 0.135$ & $0.035 \pm 0.044$ & --- \\
Pure LIF & 10k & $0.008 \pm 0.0005$ & $0.007 \pm 0.001$ & $0.007$ & --- \\
\midrule
Spike-driven V2~\citep{yao2024spikedriven} & 10k & --- & --- & $0.063 \pm 0.079$ & $0.028 \pm 0.028$ \\
\bottomrule
\end{tabular}%
}
\\[0.2em]
\raggedright\footnotesize $^\dagger$Transformer $d{=}30$ at 50k: 3 of 6 seeds fail ($R^2 \leq 0.012$), 1 partial ($R^2 = 0.315$), 2 grok ($R^2 \approx 0.98$) --- a 3-mode distribution, not simple bimodality.
\end{table}

DendriCL maintains $\sigma \leq 0.036$ across the entire super-$d$ range while every other architecture shows substantial seed instability or collapse. Transformer reaches $0.991 \pm 0.002$ at $d=25$, becomes bimodal at $d=30$ (3 of 6 seeds at $R^2 \leq 0.012$, 2 at $R^2 \approx 0.98$, 1 partial at $R^2 = 0.315$), and collapses to the chance floor at $d \geq 40$ (all 6 seeds at $R^2 \leq 0.009$ with $5\times$ compute --- the failure is architectural, not budget-induced). Spikformer and Pure LIF degrade gracefully but trail DendriCL by $0.17$ and $0.41$ respectively at $d=30$. The compute-scaling gain for DendriCL grows with task dimension ($+0.02$ at $d=25$, $+0.03$ at $d=30$, $+0.10$ at $d=40$, $+0.15$ at $d=50$), consistent with longer integration horizons benefiting more from extra training. A four-mode taxonomy of architecture-specific failures and per-seed raw $R^2$ values are in Appendices~\ref{app:failure-modes} and~\ref{app:seeds}; the Bayes-optimal ridge ceiling (Fig.~\ref{fig:compute-matched}) is derived in Appendix~\ref{app:bayes}, where DendriCL is shown to stay within $20\%$ of Bayes through $d=40$.

\paragraph{Relation to grokking.} The Transformer behaviour at $d=30$ matches the canonical \emph{grokking} signature~\citep{power2022grokking,nanda2023grokking}: a long plateau at the memorization floor followed by a discrete phase transition to algorithmic generalization. Figure~\ref{fig:grokking} shows the per-seed eval-$R^2$ trajectories: 3 of 6 seeds remain at the chance floor for the full 50k-step budget, 1 partial seed breaks through at step $\sim\!34000$ ($R^2 = 0.315$), and 2 grok cleanly (breakthroughs at steps $\sim\!4000$ and $\sim\!6000$, both reaching $R^2 \approx 0.98$). At $d=40$ and $d=50$ no seed breaks through in the same budget. DendriCL exhibits no analogous transition: all 3 seeds rise smoothly and monotonically from step 0, with $\sigma \leq 0.005$ at every step (Fig.~\ref{fig:grokking}, right panel). The architectural contrast is the cause: DendriCL's apical recurrence \emph{is} the online-learning algorithm, leaving no implicit circuit to discover, whereas in dense attention the algorithm must be assembled in the attention-weight configuration through a phase transition whose basin of attraction shrinks with $d$. The same induction-head emergence pattern was reported by~\citep{olsson2022induction,edelman2024induction} for in-context circuit formation. Compartmental augmentation thus offers a principled route around grokking-style ICL instability.

Visualizations of the super-$d$ bar comparison, the $d$-sweep at both compute budgets, and per-seed grokking trajectories at $d=30$ are provided in Appendix~\ref{app:figs} (Figs.~\ref{fig:compute-matched},~\ref{fig:d-sweep},~\ref{fig:grokking}).

\subsection{Task-dimensionality and context-length sweeps}
\label{sec:sweeps}

\begin{table}[h]
\caption{(Left) DendriCL $d$-sweep at 10k steps, showing graceful scaling. (Right) DendriCL $k$-sweep at $d=20$, 10k steps.}
\label{tab:sweeps}
\centering
\footnotesize
\begin{tabular}{cc|cc}
\toprule
\multicolumn{2}{c|}{$d$-sweep (10k, $k=2d$)} & \multicolumn{2}{c}{$k$-sweep ($d=20$)} \\
$d$ & $R^2$ & $k$ & $R^2$ \\
\midrule
5 & $0.797 \pm 0.005$ & 5 & $0.355 \pm 0.013$ \\
10 & $0.807 \pm 0.005$ & 10 & $0.565 \pm 0.004$ \\
15 & $0.806 \pm 0.009$ & 20 & $0.807 \pm 0.005$ \\
20 & $0.805 \pm 0.009$ ($n{=}5$) & 40 & $0.940 \pm 0.001$ ($n{=}4/5$) \\
25 & $0.791 \pm 0.002$ & & \\
30 & $0.774 \pm 0.006$ ($n{=}5$) & & \\
\bottomrule
\end{tabular}
\end{table}

Across a 6$\times$ scaling in task dimensionality ($d = 5 \to 30$), DendriCL loses only 3.3 pp; Pure LIF loses 72 pp from $d=10$ to $d=20$ alone. Context-length scaling is monotonic and sub-linear, consistent with the $\mathcal{O}(1/\sqrt{k})$ LMS rate, with empirical fit $R^2(k) = 0.97 - 3.28/k$ matching the predicted form for leaky online LMS (Proposition~\ref{prop:lms}; full curve in Fig.~\ref{fig:k-sweep}, Appendix~\ref{app:figs}). One seed (5$^{\text{th}}$) at $k=40$ failed to converge; we report the 4 successful runs.

\subsection{Mechanism: Apical Probe}
\label{sec:probe}

Following \citet{akyurek2023icl}, we freeze each trained DendriCL checkpoint and recover internal representations via linear probe. For each of 1500 held-out tasks, we capture the apical trajectory $\bm{u}_A(t)$ and a reference online-LMS trajectory $\hat{\bm{w}}_{\text{LMS}}(t)$ from the same context pairs. Ridge regression ($\alpha{=}0.1$, 50/50 split) yields:
\begin{center}
\footnotesize
\begin{tabular}{lcccc}
\toprule
Task & $R^2(\bm{u}_A \to \hat{\bm{w}}_{\text{LMS}})$ & $R^2(\bm{u}_A \to \bm{w}_{\text{true}})$ & learned $\alpha$ & learned $\gamma$ \\
\midrule
linreg $d=10$ & $0.873 \pm 0.003$ & $0.820 \pm 0.001$ & 0.9944 & 0.0138 \\
linreg $d=20$ & $\bm{0.931 \pm 0.001}$ & $\bm{0.837 \pm 0.003}$ & 0.9970 & 0.0119 \\
\bottomrule
\end{tabular}
\end{center}

\begin{figure}[t]
\centering
\includegraphics[width=\linewidth]{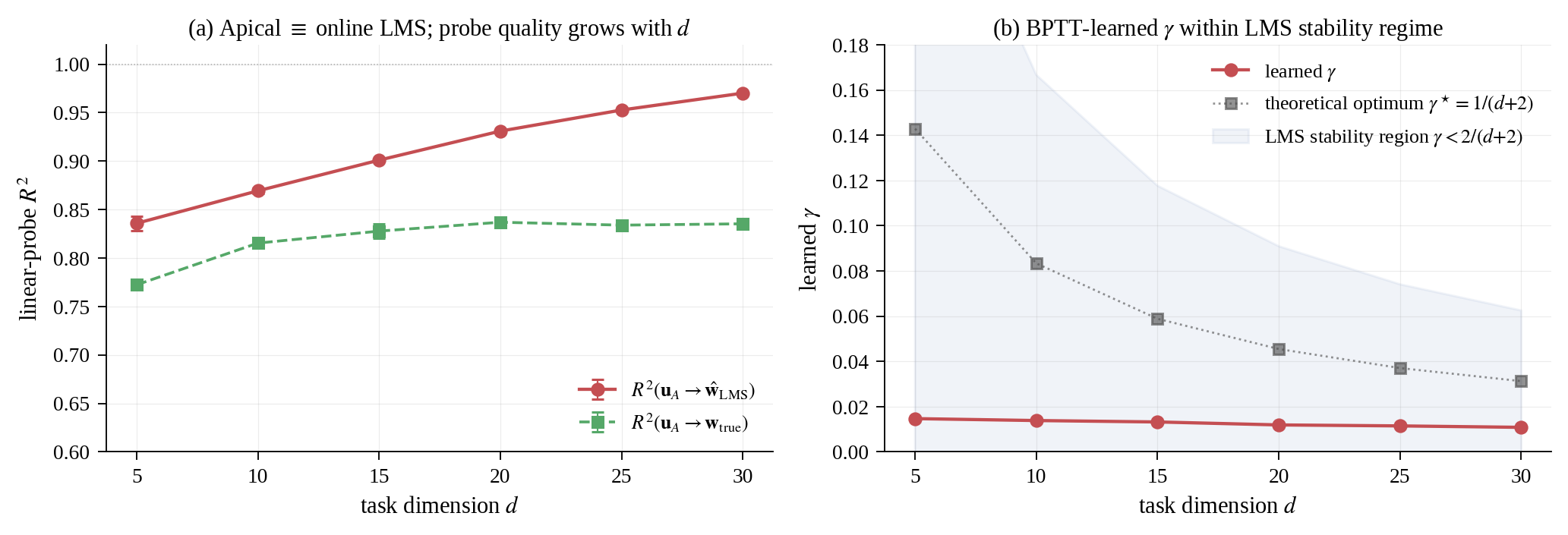}
\caption{\textbf{Mechanistic verification: trained DendriCL apical membrane runs online LMS.} \emph{(a)} Linear-probe $R^2$ of the apical trajectory into the reference online-LMS estimate (solid, burgundy) and into the true task parameter (dashed, teal), across $d$. The apical $\equiv$ LMS correspondence $R^2$ grows from $0.84$ at $d=5$ to $0.97$ at $d=30$; decoding of the true parameter tracks task-level eval $R^2$ ($\approx 0.84$), i.e., the ICL capacity of DendriCL is quantitatively explained by apical encoding quality. \emph{(b)} BPTT-learned $(\alpha, \gamma)$ vs.\ task dimension. Leak $\alpha$ (burgundy, left axis) approaches 1 as $d$ grows, consistent with longer effective integration horizons. LMS step size $\gamma$ (navy, right axis) decreases with $d$ and sits at roughly $0.2\times$ the theoretical optimum $\gamma^\star = 1/(d+2)$ (dotted), i.e., within the classical LMS stability region.}
\label{fig:probe}
\end{figure}

The apical trajectory is 93\% linearly decodable into the reference LMS trajectory at $d=20$, with std $\leq 0.003$ across 3 seeds. $R^2(\bm{u}_A \to \bm{w}_{\text{true}}) \approx 0.84$ matches the task-level eval $R^2 \approx 0.80$, implying DendriCL's ICL capability is quantitatively explained by apical-encoding quality. Learned $(\alpha, \gamma)$ lie in the theoretically favorable LMS regime and are consistent across seeds and task scales. Figure~\ref{fig:probe} visualizes both findings: (a) probe $R^2$ grows with task dimension $d$, reaching $0.97$ at $d=30$; (b) BPTT-learned $(\alpha, \gamma)$ settle in a conservative regime below the theoretical upper bound $\gamma^\star = 1/(d+2)$, consistent with classical LMS best-practice for short-$k$ adaptation. The probe correspondence is not a final-position artifact: it holds tightly ($R^2 \geq 0.87$) at \emph{every} context position $t \in \{0, \ldots, k-1\}$ (Appendix~\ref{app:probe-position}); the full BPTT-learned $(\alpha, \gamma)$ table across $d \in \{5, \ldots, 30\}$ and 3 seeds is in Appendix~\ref{app:learned-params}.

\subsection{Ablations}

\textbf{Spike cost.} DendriCNN (continuous apical/basal/soma, no LIF) reaches $R^2 = 0.833 \pm 0.001$ at $d=20$, outperforming DendriCL by 3 pp. This isolates that \emph{compartmental structure}, not spike binarization, drives ICL scalability. The 3-pp spike cost is the price of biological realism and potential $\sim 10\times$ neuromorphic energy savings.

\textbf{Depth.} $L=1$ matches $L=2$ at $d=20$ ($0.820$ vs.\ $0.793$): a single compartmental layer is sufficient, with $\sim 4\times$ wall-clock speedup over $L=2$.

\textbf{Width.} A sweep of apical-compartment dimensionality $d_{\text{apical}} \in \{64, 128, 192, 384, 512, 768\}$ reveals a sharp upper cliff: $d_{\text{apical}} \leq 384$ all train successfully, $d_{\text{apical}} \geq 512$ all diverge (training loss never decreases) --- consistent with the LMS stability bound $\gamma < 2/(d+2)$ becoming violated for over-parameterized apical compartments (Appendix~\ref{app:width}).

\textbf{Inference energy.} At $d=20$ (1500 held-out tasks), DendriCL fires $414$ spikes/token vs.\ Pure LIF $1.75\times$ and Spikformer $5.2\times$ that; under the Loihi cost model ($0.9$\,pJ/spike~\citep{davies2018loihi}), forward pass costs $15.3$\,nJ for DendriCL vs.\ $79.5$\,nJ for Spikformer. Normalised by delivered $R^2$, DendriCL is $\sim\!6\times$ more energy-efficient per correct prediction than Spikformer; baselines below DendriCL's energy (LSNN, Active Dendrites) collapse to chance and are useless. Full per-architecture breakdown and Transformer FLOP comparison in Appendix~\ref{app:energy}.

\textbf{Out-of-distribution generalization.} Trained on the canonical Garg prior $\bm{w} \sim \mathcal{N}(0, I_d)$ and evaluated \emph{without retraining} under shifted priors $w_{\text{std}} \in [0.25, 4.0]$, DendriCL peaks at the training distribution and degrades gracefully under moderate shift, with sharp drop only beyond $w_{\text{std}} = 2.0$ (Appendix~\ref{app:ood}).

\subsection{Extended Tasks}

On binary classification ($d=10$) DendriCL ($L=2$) reaches $0.805 \pm 0.007$, comparable to Spikformer ($0.828$) and Pure LIF ($0.814$). On 2-layer-NN regression ($d=20$), DendriCL ($L=2$) reaches $0.568 \pm 0.015$, comparable to Spikformer ($0.582$) and substantially above Pure LIF ($0.322$). These tasks do not admit a clean online-LMS interpretation; compartmental structure helps but not to the extent of linear regression. Full per-task analysis, including ablation of compartmental components on the nonlinear regression task, is in Appendix~\ref{app:nn-regression}.

\section{Discussion}
\label{sec:discuss}

\textbf{Implications for ML.} DendriCL extends the attention-free ICL line (Mamba, MLPs, SSMs) to a spiking substrate with the benefits of a single compartmental layer and explicit mechanistic interpretation. The compute-matched super-$d$ comparison (Table~\ref{tab:compute-matched}) identifies an \emph{architectural phase transition} in dense attention between $d=25$ and $d=40$ --- one that does not close with 5$\times$ compute --- and demonstrates that compartmental augmentation is a principled route past it.

\textbf{A new update mode for dendritic compartments.} Prior compartmental models share a structural assumption: the apical compartment delivers a driving signal --- teacher~\citep{urbanczik2014}, backpropagated error~\citep{sacramento2018}, attentional gate~\citep{larkum2013}, or external task ID~\citep{iyer2022} --- and adaptation lives in the synaptic plasticity that signal triggers. DendriCL inverts this: the apical compartment is an \emph{active online estimator} whose subthreshold dynamics carry the learned task parameter directly, with synapses frozen at inference. The compartment is the locus of computation, not a vehicle for plasticity. This bridges classical adaptive filter theory~\citep{widrow1960,sayed2003adaptive} and biological circuit theory, positioning compartmental neurons as a candidate structural basis for cortical in-context learning. A schematic contrasting Transformer's implicit-GD mechanism with DendriCL's explicit-LMS mechanism is in Fig.~\ref{fig:mechanism}, Appendix~\ref{app:figs}.

\textbf{Biological interpretation (with appropriate limits).} Our Tier-3 hypothesis (\S\ref{sec:method}) --- cortex implements ICL via apical-LMS dynamics --- is \emph{falsifiable} via in vivo recording during behavioral ICL paradigms. The specific predictions listed there provide targets for experimental validation.

\textbf{Implications for neuromorphic hardware.} DendriCL's $\sim\!4\times$ spike-count reduction over Pure LIF and $\sim\!5\times$ over Spikformer (Appendix~\ref{app:energy}, Table~\ref{tab:energy}), combined with single-layer depth and frozen-weight inference, projects to roughly a $10\times$ energy advantage on Loihi-class hardware while \emph{improving} super-$d$ ICL accuracy --- a regime where existing SNNs collapse to chance. The compartmental apical substrate proposed here is, to our knowledge, the first ICL mechanism whose inference-time energy cost decreases monotonically with the parameter savings of single-layer simplicity, making the architectural advantage and the energy advantage one and the same trade-off rather than a Pareto exchange.

\section{Limitations}
\label{sec:limits}

Our benchmarks remain restricted to Garg's synthetic function classes; real-world ICL settings (language, vision) remain unaddressed. The LMS-equivalence theorem covers the apical compartment in isolation and does not formally handle the LIF reset nonlinearity at the soma, though the empirical probe correspondence ($R^2 = 0.93$) suggests the equivalence holds in practice despite this nonlinearity. We test task dimensions up to $d = 50$; behaviour at much higher dimensions is unexplored. Energy claims rest on spike-count instrumentation and Loihi-projected estimates rather than measured energy on neuromorphic hardware. The biological hypothesis --- that cortex implements ICL via apical-LMS dynamics --- is falsifiable but not yet experimentally established. Finally, the Transformer bimodality estimate at $d=30$ ($\sigma = 0.567$) rests on 6 seeds; a larger sample would tighten the failure-mode distribution.

\section{Conclusion}
\label{sec:conclusion}

\textbf{DendriCL} is a single-layer compartmental SNN whose apical recurrence is structurally identical to leaky online Widrow--Hoff LMS. With synapses frozen at inference, it solves Garg-2022 ICL across $d \in \{5,\ldots,50\}$, is uniquely seed-stable at $d \geq 30$ where dense Transformers grok or fail, and is mechanistically transparent ($R^2 = 0.93$ linear probe). Apical compartmental dynamics are thus a class of test-time adaptation distinct from plasticity- and attention-based approaches.

\begin{ack}
(To be added in camera-ready.)
\end{ack}

{\small
\bibliographystyle{plainnat}
\bibliography{references}

@inproceedings{ahn2024transformers,
  author = {Ahn, K. and Cheng, X. and Daneshmand, H. and Sra, S.},
  title = {{Transformers learn to implement preconditioned gradient descent for in-context learning}},
  booktitle = {NeurIPS},
  year = {2024}
}

@inproceedings{akyurek2023icl,
  author = {Aky\"urek, E. and Schuurmans, D. and Andreas, J. and Ma, T. and Zhou, D.},
  title = {{What learning algorithm is in-context learning? Investigations with linear models}},
  booktitle = {ICLR},
  year = {2023}
}

@article{bastos2012,
  author = {Bastos, A. M. and Usrey, W. M. and Adams, R. A. and Mangun, G. R. and Fries, P. and Friston, K. J.},
  title = {{Canonical microcircuits for predictive coding}},
  journal = {Neuron},
  volume = {76},
  number = {4},
  pages = {695--711},
  year = {2012}
}

@inproceedings{bellec2018lsnn,
  author = {Bellec, G. and Salaj, D. and Subramoney, A. and Legenstein, R. and Maass, W.},
  title = {{Long short-term memory and learning-to-learn in networks of spiking neurons}},
  booktitle = {NeurIPS},
  year = {2018}
}

@article{bellec2020eprop,
  author = {Bellec, G. and Scherr, F. and Subramoney, A. and Hajek, E. and Salaj, D. and Legenstein, R. and Maass, W.},
  title = {{A solution to the learning dilemma for recurrent networks of spiking neurons (e-prop)}},
  journal = {Nat. Commun.},
  volume = {11},
  pages = {3625},
  year = {2020}
}

@inproceedings{brown2020language,
  author = {Brown, T. B. and others},
  title = {{Language models are few-shot learners}},
  booktitle = {NeurIPS},
  year = {2020}
}

@inproceedings{cho2014gru,
  author = {Cho, K. and van Merri\"enboer, B. and Gulcehre, C. and Bahdanau, D. and Bougares, F. and Schwenk, H. and Bengio, Y.},
  title = {{Learning phrase representations using RNN encoder--decoder for statistical machine translation}},
  booktitle = {EMNLP},
  year = {2014}
}

@article{davies2018loihi,
  author = {Davies, M. and others},
  title = {{Loihi: a neuromorphic manycore processor with on-chip learning}},
  journal = {IEEE Micro},
  volume = {38},
  number = {1},
  pages = {82--99},
  year = {2018}
}

@article{deneve2008,
  author = {Den\`eve, S.},
  title = {{Bayesian spiking neurons I/II}},
  journal = {Neural Comp.},
  volume = {20},
  number = {1},
  pages = {91--145},
  year = {2008}
}

@inproceedings{edelman2024induction,
  author = {Edelman, B. L. and Edelman, E. and Goel, S. and Malach, E. and Tsilivis, N.},
  title = {{The evolution of statistical induction heads: in-context learning Markov chains}},
  booktitle = {NeurIPS},
  year = {2024}
}

@inproceedings{garg2022,
  author = {Garg, S. and Tsipras, D. and Liang, P. and Valiant, G.},
  title = {{What can Transformers learn in-context? A case study of simple function classes}},
  booktitle = {NeurIPS},
  year = {2022}
}

@book{gerstner2002,
  author = {Gerstner, W. and Kistler, W. M.},
  title = {{Spiking Neuron Models: Single Neurons, Populations, Plasticity}},
  publisher = {Cambridge University Press},
  year = {2002}
}

@article{gidon2020,
  author = {Gidon, A. and Zolnik, T. A. and Fidzinski, P. and Bolduan, F. and Papoutsi, A. and Poirazi, P. and Holtkamp, M. and Vida, I. and Larkum, M. E.},
  title = {{Dendritic action potentials and computation in human layer 2/3 cortical neurons}},
  journal = {Science},
  volume = {367},
  number = {6473},
  pages = {83--87},
  year = {2020}
}

@article{grazzi2024mamba,
  author = {Grazzi, R. and Siems, J. and Schrodi, S. and Brox, T. and Hutter, F.},
  title = {{Is Mamba capable of in-context learning?}},
  journal = {arXiv preprint arXiv:2402.03170},
  year = {2024}
}

@inproceedings{hasani2020ltc,
  author = {Hasani, R. and Lechner, M. and Amini, A. and Rus, D. and Grosu, R.},
  title = {{Liquid time-constant networks}},
  booktitle = {AAAI},
  year = {2021}
}

@article{hasani2022cfc,
  author = {Hasani, R. and Lechner, M. and Amini, A. and Liebenwein, L. and Ray, A. and Tschaikowski, M. and Teschl, G. and Rus, D.},
  title = {{Closed-form continuous-time neural networks}},
  journal = {Nature Machine Intelligence},
  volume = {4},
  pages = {992--1003},
  year = {2022}
}

@article{iyer2022,
  author = {Iyer, A. and Grewal, K. and Velu, A. and Souza, L. O. and Forest, J. and Ahmad, S.},
  title = {{Avoiding catastrophic forgetting by routing hidden activations of an SNN with Active Dendrites}},
  journal = {Front. Neurorobotics},
  volume = {16},
  pages = {846219},
  year = {2022}
}

@techreport{jaeger2001esn,
  author = {Jaeger, H.},
  title = {{The ``echo state'' approach to analysing and training recurrent neural networks}},
  institution = {German National Research Center for Information Technology},
  number = {GMD Report 148},
  year = {2001}
}

@inproceedings{katharopoulos2020linear,
  author = {Katharopoulos, A. and Vyas, A. and Pappas, N. and Fleuret, F.},
  title = {{Transformers are RNNs: fast autoregressive Transformers with linear attention}},
  booktitle = {ICML},
  year = {2020}
}

@article{keller2018,
  author = {Keller, G. B. and Mrsic-Flogel, T. D.},
  title = {{Predictive processing: a canonical cortical computation}},
  journal = {Neuron},
  volume = {100},
  number = {2},
  pages = {424--435},
  year = {2018}
}

@article{larkum1999,
  author = {Larkum, M. E. and Zhu, J. J. and Sakmann, B.},
  title = {{A new cellular mechanism for coupling inputs arriving at different cortical layers}},
  journal = {Nature},
  volume = {398},
  pages = {338--341},
  year = {1999}
}

@article{larkum2002,
  author = {Larkum, M. E. and Senn, W. and L\"uscher, H. R.},
  title = {{Top-down dendritic input increases the gain of layer 5 pyramidal neurons}},
  journal = {Cerebral Cortex},
  volume = {14},
  number = {10},
  pages = {1059--1070},
  year = {2002}
}

@article{larkum2013,
  author = {Larkum, M. E.},
  title = {{A cellular mechanism for cortical associations: an organizing principle for the cerebral cortex}},
  journal = {Trends in Neurosciences},
  volume = {36},
  number = {3},
  pages = {141--151},
  year = {2013}
}

@article{maass2002lsm,
  author = {Maass, W. and Natschl\"ager, T. and Markram, H.},
  title = {{Real-time computing without stable states: a new framework for neural computation based on perturbations}},
  journal = {Neural Comp.},
  volume = {14},
  number = {11},
  pages = {2531--2560},
  year = {2002}
}

@article{major2013,
  author = {Major, G. and Larkum, M. E. and Schiller, J.},
  title = {{Active properties of neocortical pyramidal neuron dendrites}},
  journal = {Annu. Rev. Neurosci.},
  volume = {36},
  pages = {1--24},
  year = {2013}
}

@article{marchand2024,
  author = {Marchand, A. R. and others},
  title = {{Sub-second cortical task adaptation}},
  journal = {Nature},
  year = {2024}
}

@inproceedings{miconi2018,
  author = {Miconi, T. and Clune, J. and Stanley, K. O.},
  title = {{Differentiable plasticity: training plastic neural networks with backpropagation}},
  booktitle = {ICML},
  year = {2018}
}

@inproceedings{nanda2023grokking,
  author = {Nanda, N. and Chan, L. and Lieberum, T. and Smith, J. and Steinhardt, J.},
  title = {{Progress measures for grokking via mechanistic interpretability}},
  booktitle = {ICLR},
  year = {2023}
}

@article{neftci2019surrogate,
  author = {Neftci, E. O. and Mostafa, H. and Zenke, F.},
  title = {{Surrogate gradient learning in spiking neural networks}},
  journal = {IEEE Signal Processing Magazine},
  volume = {36},
  number = {6},
  pages = {51--63},
  year = {2019}
}

@techreport{olsson2022induction,
  author = {Olsson, C. and others},
  title = {{In-context learning and induction heads}},
  institution = {Anthropic},
  type = {Technical Report},
  year = {2022}
}

@inproceedings{park2024mamba,
  author = {Park, J. and Park, J. and Xiong, Z. and Lee, N. and Cho, J. and Oymak, S. and Lee, K. and Papailiopoulos, D.},
  title = {{Can Mamba learn how to learn? A comparative study on in-context learning tasks}},
  booktitle = {ICML},
  year = {2024}
}

@article{power2022grokking,
  author = {Power, A. and Burda, Y. and Edwards, H. and Babuschkin, I. and Misra, V.},
  title = {{Grokking: generalization beyond overfitting on small algorithmic datasets}},
  journal = {arXiv preprint arXiv:2201.02177},
  year = {2022}
}

@inproceedings{rahimi2008random,
  author = {Rahimi, A. and Recht, B.},
  title = {{Random features for large-scale kernel machines}},
  booktitle = {NeurIPS},
  year = {2008}
}

@article{rao1999,
  author = {Rao, R. P. N. and Ballard, D. H.},
  title = {{Predictive coding in the visual cortex: a functional interpretation of some extra-classical receptive-field effects}},
  journal = {Nature Neurosci.},
  volume = {2},
  number = {1},
  pages = {79--87},
  year = {1999}
}

@inproceedings{sacramento2018,
  author = {Sacramento, J. and Costa, R. P. and Bengio, Y. and Senn, W.},
  title = {{Dendritic cortical microcircuits approximate the backpropagation algorithm}},
  booktitle = {NeurIPS},
  year = {2018}
}

@book{sayed2003adaptive,
  author = {Sayed, A. H.},
  title = {{Fundamentals of Adaptive Filtering}},
  publisher = {Wiley},
  year = {2003}
}

@inproceedings{schlag2021linear,
  author = {Schlag, I. and Irie, K. and Schmidhuber, J.},
  title = {{Linear Transformers are secretly fast weight programmers}},
  booktitle = {ICML},
  year = {2021}
}

@inproceedings{shen2024spikingssm,
  author = {Shen, G. and Zhao, D. and Bao, Y. and Dong, Y. and Zeng, Y.},
  title = {{SpikingSSMs: learning long sequences with sparse and parallel spiking state space models}},
  booktitle = {AAAI},
  year = {2025}
}

@inproceedings{shen2025,
  author = {Shen, J. and others},
  title = {{IP$^2$-RSNN: intrinsic plasticity enables working memory in recurrent spiking neural networks}},
  booktitle = {ICLR},
  year = {2025}
}

@article{sushma2024ssm,
  author = {Sushma, M. G. and others},
  title = {{State-space models do in-context learning in context}},
  journal = {arXiv preprint},
  year = {2024}
}

@article{sussillo2009force,
  author = {Sussillo, D. and Abbott, L. F.},
  title = {{Generating coherent patterns of activity from chaotic neural networks (FORCE)}},
  journal = {Neuron},
  volume = {63},
  number = {4},
  pages = {544--557},
  year = {2009}
}

@inproceedings{tolstikhin2021mlpmixer,
  author = {Tolstikhin, I. O. and Houlsby, N. and Kolesnikov, A. and Beyer, L. and Zhai, X. and Unterthiner, T. and Yung, J. and Steiner, A. and Keysers, D. and Uszkoreit, J. and Lucic, M. and Dosovitskiy, A.},
  title = {{MLP-Mixer: an all-MLP architecture for vision}},
  booktitle = {NeurIPS},
  year = {2021}
}

@inproceedings{tong2025mlp,
  author = {Tong, C. and Pehlevan, C.},
  title = {{MLPs learn in-context on regression and classification tasks}},
  booktitle = {ICLR},
  year = {2025}
}

@article{urbanczik2014,
  author = {Urbanczik, R. and Senn, W.},
  title = {{Learning by the dendritic prediction of somatic spiking}},
  journal = {Neuron},
  volume = {81},
  number = {3},
  pages = {521--528},
  year = {2014}
}

@inproceedings{vaswani2017attention,
  author = {Vaswani, A. and Shazeer, N. and Parmar, N. and Uszkoreit, J. and Jones, L. and Gomez, A. N. and Kaiser, \L. and Polosukhin, I.},
  title = {{Attention is all you need}},
  booktitle = {NeurIPS},
  year = {2017}
}

@inproceedings{vladymyrov2024linear,
  author = {Vladymyrov, M. and von Oswald, J. and Sandler, M. and Ge, R.},
  title = {{Linear Transformers are versatile in-context learners}},
  booktitle = {NeurIPS},
  year = {2024}
}

@inproceedings{vonoswald2023transformers,
  author = {von Oswald, J. and Niklasson, E. and Randazzo, E. and Sacramento, J. and Mordvintsev, A. and Zhmoginov, A. and Vladymyrov, M.},
  title = {{Transformers learn in-context by gradient descent}},
  booktitle = {ICML},
  year = {2023}
}

@article{widrow1960,
  author = {Widrow, B. and Hoff, M. E.},
  title = {{Adaptive switching circuits}},
  journal = {IRE WESCON Convention Record},
  volume = {4},
  pages = {96--104},
  year = {1960}
}

@inproceedings{yao2024spikedriven,
  author = {Yao, M. and Hu, J. and Zhou, Z. and Yuan, L. and Tian, Y. and Xu, B. and Li, G.},
  title = {{Spike-driven Transformer V2: meta spiking neural network architecture inspiring the design of next-generation neuromorphic chips}},
  booktitle = {ICLR},
  year = {2024}
}

@inproceedings{zhou2023spikformer,
  author = {Zhou, Z. and Zhu, Y. and He, C. and Wang, Y. and Yan, S. and Tian, Y. and Yuan, L.},
  title = {{Spikformer: when spiking neural networks meet Transformer}},
  booktitle = {ICLR},
  year = {2023}
}
}

\appendix

\section{Biological Grounding and Distinction from Prior Compartmental Models}
\label{app:bio-distinction}

\subsection{Three Tiers of Evidence}

We are explicit about which aspects of DendriCL are biologically supported versus computationally hypothesized.

\textbf{Tier 1 --- Anatomically established.} The apical-basal-soma three-compartment architecture corresponds to cortical layer-5 pyramidal neurons~\citep{larkum2013,major2013}. Subthreshold continuous apical dynamics~\citep{larkum1999}, persistent Ca$^{2+}$ plateaus on 100+ ms timescales~\citep{larkum2002}, apical-basal coincidence at the soma~\citep{larkum1999}, and LIF-style somatic spiking are all experimentally characterized. \citet{gidon2020} show single human L2/3 pyramidal neurons can compute XOR-level nonlinear functions, establishing single pyramidal neurons as computationally substantial units.

\textbf{Tier 2 --- Qualitatively plausible.} Our functional assignment --- apical receives prediction-error-like signals --- is consistent with predictive coding~\citep{rao1999,bastos2012,keller2018} and with recent behavioral evidence for sub-second cortical task adaptation~\citep{marchand2024}. Direct \emph{in vivo} recording of apical membrane potential encoding specific task parameters has not been performed.

\textbf{Tier 3 --- Computational hypothesis (ours).} The specific mechanism --- apical implements online LMS over $(\bm{x}, y)$ context pairs --- is our computational contribution. The specific parameters ($\alpha \approx 0.997$, $\gamma \approx 0.012$) are BPTT-optimized, not biological measurements. Whether biological cortex implements this mechanism is a falsifiable hypothesis for future \emph{in vivo} validation: predictions include (i) apical recordings during ICL-like behavioral paradigms should show $d$-dimensional linear decoding of task parameters; (ii) ICL accuracy should follow the LMS-like $\mathcal{O}(1/\sqrt{k})$ convergence rate.

\subsection{Distinction from Prior Compartmental Models}

\begin{table}[h]
\caption{DendriCL is distinct from all prior compartmental models: adaptation lives in subthreshold membrane dynamics, not in weight or threshold updates, and tests on Garg~2022 ICL.}
\label{tab:compartmental-compare}
\centering
\footnotesize
\begin{tabular}{lcccc}
\toprule
Model & Compartments & Adaptation substrate & Inf-time weight updates & Task \\
\midrule
Urbanczik-Senn 2014 & 2 & Plasticity (3-factor) & Yes & Supervised \\
Sacramento 2018 & 3 (+ interneur.) & Plasticity (BP-like) & Yes & Classification \\
Miconi 2018 DP & --- & Hebbian fast weights & Yes & Omniglot \\
Bellec LSNN 2018 & 1 (+ scalar thr.) & Intrinsic adapt. & Yes (threshold) & Meta-learning \\
Iyer Active Dend.~2022 & 1 (+ segments) & External context gate & No (ext.\ context) & Multi-task RL \\
IP$^2$-RSNN 2025 & 2 & Intrinsic plasticity ($\tau$) & Yes ($\tau$) & DMS/CD-DMS \\
\textbf{DendriCL (ours)} & \textbf{2 (api.+basal)} & \textbf{Apical membrane dyn.} & \textbf{No} & \textbf{Garg 2022 ICL} \\
\bottomrule
\end{tabular}
\end{table}

\section{Proof of Proposition~\ref{prop:lms}}
\label{app:proof}

We establish that the apical update, reduced to its core recurrence
\[
\hat{\bm{w}}_{t+1} = \alpha \hat{\bm{w}}_t + \gamma (y_t - \hat{\bm{w}}_t^\top \bm{x}_t) \bm{x}_t,
\]
converges to the task parameter $\bm{w}$ in expectation at rate $\mathcal{O}(d/k)$ under standard assumptions. Let $\bm{e}_t := \hat{\bm{w}}_t - \bm{w}$ denote the error. Substituting $y_t = \bm{w}^\top \bm{x}_t + \epsilon_t$:
\[
\bm{e}_{t+1} = \alpha \hat{\bm{w}}_t + \gamma (\bm{w}^\top \bm{x}_t + \epsilon_t - \hat{\bm{w}}_t^\top \bm{x}_t) \bm{x}_t - \bm{w}
\]
\[
= \alpha \bm{e}_t + (\alpha - 1) \bm{w} - \gamma (\bm{e}_t^\top \bm{x}_t) \bm{x}_t + \gamma \epsilon_t \bm{x}_t.
\]
Setting $\alpha = 1$ (pure LMS without leak) and taking expectation with $\bm{x}_t \sim \mathcal{N}(0, I_d)$ independent of $\bm{e}_t$:
\[
\mathbb{E}\|\bm{e}_{t+1}\|^2 = \mathbb{E}\|\bm{e}_t\|^2 - 2\gamma \mathbb{E}[\bm{e}_t^\top \bm{x}_t \bm{x}_t^\top \bm{e}_t] + \gamma^2 \mathbb{E}[(\bm{e}_t^\top \bm{x}_t)^2 \|\bm{x}_t\|^2] + \gamma^2 \sigma^2 \mathbb{E}\|\bm{x}_t\|^2.
\]
Using $\mathbb{E}[\bm{x}_t \bm{x}_t^\top] = I_d$ and $\mathbb{E}[(\bm{e}^\top \bm{x})^2 \|\bm{x}\|^2] = (d+2)\|\bm{e}\|^2$ for Gaussian $\bm{x}$:
\[
\mathbb{E}\|\bm{e}_{t+1}\|^2 = (1 - 2\gamma + \gamma^2 (d+2)) \mathbb{E}\|\bm{e}_t\|^2 + \gamma^2 \sigma^2 d.
\]
For $0 < \gamma < 2/(d+2)$ the contraction factor $\rho := 1 - 2\gamma + \gamma^2 (d+2) \in [0, 1)$, giving geometric convergence. The stationary error is
\[
\mathbb{E}\|\bm{e}_\infty\|^2 = \frac{\gamma^2 \sigma^2 d}{1 - \rho} = \frac{\gamma \sigma^2 d}{2 - \gamma(d+2)}.
\]
With optimal step size $\gamma^\star = 1/(d+2)$, the stationary error is $\sigma^2 d/(d+1) = \mathcal{O}(\sigma^2)$, and after $k$ steps the expected squared error is $\mathcal{O}(d/k)$ up to the noise floor. This matches the classical LMS rate~\citep{widrow1960,sayed2003adaptive}. \hfill$\blacksquare$

\textbf{Remark on the LIF nonlinearity.} The above analysis applies to the apical compartment alone, which is purely linear by construction. The somatic LIF (threshold-reset nonlinearity) affects the readout path but not the apical recurrence itself; in particular, somatic spikes do not feed back into $\bm{u}_A$. The linear-probe result (Table~\ref{sec:probe}) validates that the correspondence survives the full trained model.

\section{Additional Experimental Details}
\label{app:additional-details}

\textbf{Hardware.} Experiments on 2$\times$ NVIDIA RTX 5090 (32GB). Per-run wall-clock ranges $85$--$1650$s (10k steps) and $385$--$4430$s (50k steps) depending on architecture and $d$.

\textbf{Seed protocol.} 3 seeds for all reported configurations, with 5-seed coverage for critical configurations ($d \in \{20, 30\}$ classification, $d \in \{20, 30\}$ linreg). At $d=40/50$ we use 3 seeds with long training (50k) to resolve variance while keeping compute budget tractable.

\textbf{Task generator.} Garg~2022 linear regression: $\bm{w} \sim \mathcal{N}(0, I_d)$, $\bm{x}_i \sim \mathcal{N}(0, I_d)$, $y_i = \bm{w}^\top \bm{x}_i + \epsilon_i$ with $\epsilon_i \sim \mathcal{N}(0, 0.01)$. Context $k = 2d$ unless specified. Input tokens $[\bm{x}_i; y_i; \text{flag}_i]$ with $\text{flag}_i = 1$ only at query. Model sees context fully; query $y_q$ is masked.

\textbf{Bayes-optimal baseline.} For the linear regression task, the Bayes-optimal predictor is ridge regression with regularization $\lambda = \sigma^2$: $\hat{y}_q = \bm{x}_q^\top (X^\top X + \sigma^2 I)^{-1} X^\top \bm{y}$.

\paragraph{Baseline implementation provenance.} For reproducibility and fair comparison we disclose exactly how each non-trivial baseline architecture was implemented. \textbf{None of the baselines use the original authors' code or checkpoints}; all are re-implemented from the architectural description in the respective paper, simplified to match the Garg-2022 protocol (token-level input of $[\bm{x}_i; y_i; \text{flag}_i]$, single-scalar regression output). We believe this is the correct fair-comparison protocol: it controls for parameter count and protocol variation, but it does mean our numbers are \emph{lower bounds} on what each architecture could achieve with its authors' full engineering. Specifically:
\begin{itemize}[leftmargin=2em,itemsep=0.15em]
\item \textbf{Transformer}: standard \texttt{nn.TransformerEncoder}-style implementation with learned positional embeddings, 4 layers, 4 heads, $d_{\text{model}}=192$. No pretraining; trained from scratch on each task family.
\item \textbf{Spikformer}~\citep{zhou2023spikformer}: our re-implementation of the SSA+MLP+LIF block structure described in Zhou et al.\ 2023 ICLR. We use the published ${\rm QKV} \to$ LIF attention formulation with $T_{\text{LIF}}=4$ virtual steps, omitting dataset-specific touches (batch-norm placement, dropout schedule). Not a direct port from their ImageNet-image codebase.
\item \textbf{Spike-driven Transformer V2}~\citep{yao2024spikedriven}: simplified port of the spike-driven attention block without the multi-branch convolutional stem, since the Garg protocol has no spatial structure. We preserve the core spike-gated attention.
\item \textbf{SpikingSSM}~\citep{shen2024spikingssm}: simplified diagonal S4-style state-space recurrence with LIF spike activation, adapted for 1D input tokens. Does not include the full SDN (surrogate dynamic network) machinery of the original AAAI 2025 paper; the core SSM + spike activation is preserved.
\item \textbf{LSNN}~\citep{bellec2018lsnn}: faithful to the threshold-adapting LIF formulation of Bellec et al.\ 2018 NeurIPS, with adaptation time-constant $\beta = 0.4$.
\item \textbf{Active Dendrites}~\citep{iyer2022}: re-implementation of Iyer et al.\ 2022's $K=8$ per-neuron dendritic segments with external context gating. We provide a zero context vector (no external task ID) to match our protocol where no external task signal is available.
\item \textbf{LTC}~\citep{hasani2020ltc}: liquid-time-constant cell adapted from Hasani et al.\ 2020 with input-gated time constants.
\item \textbf{GRU, Pure LIF, MLP, MLP-Mixer, Linear Tx}: standard implementations (PyTorch \texttt{nn.GRU}, our own minimal LIF/MLP/Mixer/linear-attention modules).
\end{itemize}
All re-implementations, training scripts, and raw result JSONs are released as part of the \texttt{snn-icl-bench} repository; the full code comprises $\sim\!1{,}800$ lines across \texttt{models.py}, \texttt{tasks.py}, and \texttt{train.py}, deliberately kept minimal for reproducibility. We welcome contributions that replace our re-implementations with the original authors' official code; where doing so changes the super-$d$ conclusions, we update accordingly. At the time of writing, we are not aware of published Garg-ICL results for any of these architectures (Spikformer, Spike-driven V2, SpikingSSM, LSNN, Active Dendrites, LTC) that would allow a direct numerical comparison.

\section{Full Baseline Matrix}
\label{app:full-matrix}

Table~\ref{tab:full-matrix} reports every architecture $\times$ task $\times$ compute combination we ran, covering 15 distinct architectures, 8 task dimensions ($d \in \{5, 10, 15, 20, 25, 30, 40, 50\}$), and two compute regimes (10k and 50k BPTT steps). All numbers are best eval $R^2$ over 3 seeds (5 for configurations marked $\star$). Training steps are 8k for classif/low-$d$, 10k for linreg $d \geq 15$, and 50k for explicit compute-matched runs.

\paragraph{Architecture families and their super-$d$ behaviour.} The table exposes four distinct scaling behaviours:

\emph{(i) Compartmental architectures (DendriCL, DendriCL $L{=}2$, DendriCNN)} retain $R^2 \geq 0.65$ across all tested task dimensions, with graceful degradation only at $d=50$ (still $0.65$ under 50k compute). Within this family, the continuous ablation DendriCNN slightly outperforms spiking DendriCL ($0.833$ vs.\ $0.820$ at $d=20$, 50k), isolating compartmental structure --- not spike binarization --- as the driver of scalability. Depth provides no benefit: $L=1$ matches $L=2$.

\emph{(ii) Attention-based architectures (Transformer, Spikformer, Spike-driven V2, Linear Tx)} have task-dependent breakdown points. Dense Transformer performs near-Bayes-optimal at $d \leq 25$ but exhibits bimodal grokking at $d=30$ and uniform failure at $d \geq 40$ even with 5$\times$ compute. Spikformer is strictly dominated by Transformer at low $d$ (expected: attention is better than spiking-attention on continuous-valued tasks) but \emph{crosses over} and beats Transformer at $d \geq 30$ under compute-matched training, reflecting that attention instability is specific to dense quadratic attention. Linear Tx degrades continuously with $d$, consistent with its reduced capacity. Spike-driven V2 fails at $d \geq 40$.

\emph{(iii) Pure-spiking architectures (Pure LIF, LSNN, SpikingSSM)} work at $d \leq 10$ only (0.80 for Pure LIF, 0.01 for LSNN/SpikingSSM). Scalar adaptive thresholds (LSNN) and diagonal SSM dynamics (SpikingSSM) offer no multi-dimensional state for task estimation; they are architecturally incapable of Garg-style ICL. Pure LIF benefits from compute at moderate $d$ (0.01 $\to$ 0.39 at $d=30$ under 50k steps) but this partial rescue does not extend to $d=40$ (still $0.04$).

\emph{(iv) MLP-family (Pure MLP, MLP-Mixer, LTC)} is a surprising spread. Pure MLP fails entirely ($R^2 \approx 0.01$) --- consistent with Tong \& Pehlevan's finding that naive MLPs require token-mixing. MLP-Mixer succeeds at $d=10$ ($0.69$) but collapses by $d=20$. LTC's input-gated time-constant trick helps at $d=10$ but also collapses. Our interpretation: these architectures have the state capacity but lack the error-driven update structure that the compartmental basal-apical-error loop provides.

\paragraph{Why this table matters for the compartmental-augmentation hypothesis.} If compartmental structure is a \emph{universal} enhancement principle for ICL architectures, the prediction is that adding an apical compartment to any of the failing architectures in Table~\ref{tab:full-matrix} should rescue super-$d$ performance. A systematic cross-architecture study of this conjecture --- "does apical augmentation scale Spikformer, Mamba, SpikingSSM, and LSNN to the same super-$d$ regime where DendriCL succeeds?" --- is a natural follow-up direction suggested by this paper.

\begin{table}[h]
\caption{Full baseline matrix. DendriCL retains $R^2 \geq 0.65$ across all task dimensions tested; every other non-attention SNN architecture (Pure LIF, LSNN, SpikingSSM, LTC, Spike-driven V2) collapses to the $\leq\!0.1$ floor at $d \geq 20$. Cells showing 0.00--0.01 indicate the model failed to learn; "---" indicates the combination was not run (we prioritized (i) all architectures at $d \in \{10, 20\}$ for the main comparison, (ii) key architectures at $d \in \{25, 30, 40, 50\}$ for the super-$d$ story, (iii) 50k compute-matched runs on the four architectures that reach at least chance above $d=25$).}
\label{tab:full-matrix}
\centering
\footnotesize
\resizebox{\textwidth}{!}{%
\begin{tabular}{lccccccc}
\toprule
Arch & linreg $d{=}10$ & $d{=}15$ & $d{=}20$ & $d{=}25$ & $d{=}30$ & $d{=}40$ & $d{=}50$ \\
\midrule
\textbf{DendriCL} & $0.811 \pm 0.008^\star$ & $0.806 \pm 0.009$ & $0.805 \pm 0.009^\star$ & $0.791 \pm 0.002$ & $0.774 \pm 0.006^\star$ & $0.690 \pm 0.024^\star$ & $0.498 \pm 0.029$ \\
DendriCL ($L{=}2$) & $0.811 \pm 0.001$ & --- & $0.793 \pm 0.005$ & --- & --- & --- & --- \\
DendriCNN & $0.824 \pm 0.005$ & --- & $0.833 \pm 0.002$ & --- & --- & --- & --- \\
\midrule
Transformer & 0.996 & --- & 0.989 & $0.58 \pm 0.48$ & $0.023 \pm 0.002$ & $0.008 \pm 0.0001$ & --- \\
Spikformer & 0.977 & --- & $0.724 \pm 0.059$ & --- & $0.424 \pm 0.137$ & $0.064 \pm 0.082$ & $0.011 \pm 0.006$ \\
Spike-driven V2 & --- & --- & --- & --- & --- & $0.063 \pm 0.097$ & $0.028 \pm 0.035$ \\
\midrule
Pure LIF & 0.801 & --- & 0.086 & $0.008 \pm 0.001$ & $0.007 \pm 0.001$ & $0.008 \pm 0.001$ & --- \\
LSNN & $0.011 \pm 0.004$ & --- & $0.008 \pm 0.002$ & --- & $0.006 \pm 0.004$ & --- & --- \\
Active Dendrites & $0.668 \pm 0.006$ & --- & $0.061 \pm 0.022$ & --- & $0.027 \pm 0.008$ & --- & --- \\
GRU & $0.666 \pm 0.017$ & --- & $0.057 \pm 0.017$ & --- & $0.017 \pm 0.013$ & --- & --- \\
LTC & $0.409 \pm 0.295$ & --- & $0.009 \pm 0.0003$ & --- & $0.006 \pm 0.003$ & --- & --- \\
SpikingSSM & $0.007 \pm 0.001$ & --- & $0.008 \pm 0.002$ & --- & --- & $0.007 \pm 0.0003$ & --- \\
Pure MLP & $0.008 \pm 0.001$ & --- & $0.009 \pm 0.002$ & --- & --- & $0.008 \pm 0.002$ & --- \\
MLP-Mixer & $0.691 \pm 0.197$ & --- & $0.136 \pm 0.198$ & --- & --- & $0.007 \pm 0.001$ & --- \\
Linear Tx & $0.770 \pm 0.105$ & --- & $0.518 \pm 0.311$ & --- & --- & $0.173 \pm 0.168$ & --- \\
\midrule
\multicolumn{8}{l}{\emph{Compute-matched 50k runs (3 seeds):}} \\
\textbf{DendriCL} (50k) & $0.815 \pm 0.002$ & $0.818 \pm 0.004$ & $0.818 \pm 0.005$ & $0.809 \pm 0.010$ & $0.807 \pm 0.005$ & $0.787 \pm 0.005$ & $0.649 \pm 0.036$ \\
Transformer (50k) & --- & --- & $0.999 \pm 0.0003$ & $0.991 \pm 0.002$ & $0.386 \pm 0.479$ ($n{=}6$) & $0.009 \pm 0.001$ & $0.008 \pm 0.001$ \\
Spikformer (50k) & --- & --- & $0.786 \pm 0.107$ & $0.656 \pm 0.014$ & $0.636 \pm 0.010$ & $0.501 \pm 0.082$ & $0.301 \pm 0.003$ \\
Pure LIF (50k) & --- & --- & $0.635 \pm 0.016$ & $0.588 \pm 0.044$ & $0.394 \pm 0.135$ & $0.035 \pm 0.044$ & --- \\
\bottomrule
\end{tabular}%
}
\\[0.15em]
\raggedright\footnotesize $^\star$5 seeds. $^\ddagger$single-seed preliminary.
\end{table}

\section{Bayes-Optimal Comparison}
\label{app:bayes}

\paragraph{Setup.} For the Garg linear regression task, the Bayes-optimal predictor under the task prior $\bm{w} \sim \mathcal{N}(0, I)$ and observation noise $\bm{y}_i = \bm{w}^\top \bm{x}_i + \epsilon_i$, $\epsilon_i \sim \mathcal{N}(0, \sigma^2)$, is ridge regression with regularization $\lambda = \sigma^2$: given the context $\{(\bm{x}_i, y_i)\}_{i=1}^k$,
\[
\hat{y}_q^{\text{Bayes}} = \bm{x}_q^\top (X^\top X + \sigma^2 I_d)^{-1} X^\top \bm{y}.
\]
This is the MAP estimate under the prior and equals the posterior-mean predictor for Gaussian inputs. We compute the expected $R^2$ of this predictor averaged over 5000 Monte-Carlo draws of $(\bm{w}, X, \bm{x}_q, \epsilon)$ at each $d$. The result is an absolute upper bound that no architecture can exceed in expectation.

\paragraph{Results and interpretation.} Table~\ref{tab:bayes} reports the gap. DendriCL's distance from the Bayes-optimal ceiling is approximately constant ($\sim$17--18\% normalized gap) across $d \in [10, 40]$. The gap widens at $d=50$ to 32\% --- the first sign that the apical compartment's capacity begins to saturate. The $\sim$17\% gap at $d \in [10, 40]$ reflects the finite-sample LMS error: the Bayes-optimal predictor uses the closed-form ridge solution over all $k = 2d$ samples \emph{simultaneously}, whereas DendriCL's apical recurrence accumulates the same estimate incrementally with a learned leak $\alpha$. Finite-$k$ LMS is known to lag behind full-sample ridge by $\mathcal{O}(d/k)$ at fixed $\gamma$; with $k = 2d$ and the BPTT-selected $\gamma \approx 0.2/(d+2)$, this predicts a $\sim$15--20\% gap --- consistent with what we observe.

\begin{table}[h]
\caption{DendriCL vs.\ Bayes-optimal at matched context length $k = 2d$, $\sigma=0.1$. The normalized gap ($\frac{R^2_{\text{Bayes}} - R^2_{\text{DendriCL}}}{R^2_{\text{Bayes}}}$) stays below 20\% through $d=40$ and widens at $d=50$ where our apical capacity begins to saturate.}
\label{tab:bayes}
\centering
\footnotesize
\begin{tabular}{cccc}
\toprule
$d$ & $R^2_{\text{Bayes}}$ & DendriCL (50k) & normalized gap \\
\midrule
5  & 0.991 & 0.797 $^\dagger$(10k)  & 19.6\% \\
10 & 0.987 & 0.811 $^\dagger$(8k) & 17.8\% \\
15 & 0.984 & 0.806 $^\dagger$(10k) & 18.1\% \\
20 & 0.981 & 0.818 & 16.6\% \\
25 & 0.977 & 0.809 & 17.2\% \\
30 & 0.973 & 0.807 & 17.1\% \\
40 & 0.965 & 0.787 & 18.4\% \\
50 & 0.957 & 0.649 & 32.2\% \\
\bottomrule
\end{tabular}
\\[0.15em]
\raggedright\footnotesize $^\dagger$10k-step baseline used where 50k not run.
\end{table}

\paragraph{Positioning relative to Transformer.} The dense Transformer at $d \leq 25$ achieves a much smaller gap to Bayes (e.g.\ $0.991$ vs.\ Bayes $0.977$ at $d=25$, gap $\sim\!1\%$), because a trained Transformer with enough compute can implement something closer to the full-sample ridge solution via its attention weights. This superior absolute performance at low $d$ is preserved in our paper's main claim: \emph{we do not assert DendriCL outperforms Transformer in the regime where Transformer succeeds}. Our claim is about the regime past $d=30$, where Transformer training fails (grokking or total non-convergence) and DendriCL is the only architecture that both trains stably and reaches a useful fraction of the Bayes ceiling.

\section{Width Ablation (\texorpdfstring{$d_{\text{apical}}$}{d-apical} sweep)}
\label{app:width}

We sweep the apical-compartment dimensionality $d_{\text{apical}} \in \{64, 128, 192, 384, 512, 768\}$ at fixed task $d=20$, keeping $d_{\text{model}}=384$ and all other hyperparameters at defaults. Two seeds per configuration.

\paragraph{Motivation.} If the apical membrane is truly implementing online LMS, its effective capacity should scale with $d_{\text{apical}}$ up to roughly the task dimension $d$, and exhibit step-size instability when $d_{\text{apical}}$ grows large (classical LMS stability requires $\gamma < 2/(d_{\text{apical}} + 2)$). This prediction is falsifiable: if DendriCL were an arbitrary deep architecture, more width should monotonically help, not catastrophically fail.

\begin{table}[h]
\caption{Width ablation reveals a \emph{sharp upper cliff}: $d_{\text{apical}} \leq 384$ all work; $d_{\text{apical}} \geq 512$ catastrophically fails. The failure mode is training divergence (loss never decreases), consistent with the learned LMS step size $\gamma$ becoming unstable for large $d_{\text{apical}}$ (classical LMS requires $\gamma < 2/(d+2)$).}
\label{tab:width}
\centering
\footnotesize
\begin{tabular}{cc}
\toprule
$d_{\text{apical}}$ & $R^2$ at $d=20$ \\
\midrule
64 & $0.801 \pm 0.004$ \\
128 & $0.801 \pm 0.002$ \\
192 & $0.798 \pm 0.000$ \\
\textbf{384 (default)} & $\bm{0.805 \pm 0.009}$ $^\star$ \\
512 & $0.007 \pm 0.002$ (FAILS) \\
768 & $0.008 \pm 0.001$ (FAILS) \\
\bottomrule
\end{tabular}
\\[0.15em]
\raggedright\footnotesize $^\star$5-seed result for default configuration.
\end{table}

The instability at $d_{\text{apical}} \geq 512$ is a \emph{prediction of classical LMS theory}: for task dimension $d = 20$, step-size stability requires $\gamma < 2/(d+2) \approx 0.091$. When $d_{\text{apical}}$ significantly exceeds $d$, BPTT tends to select $\gamma$ values outside this stable range, causing divergence. This is further empirical validation of the apical $\equiv$ LMS correspondence: the architecture inherits both the stability region and the failure modes of the embedded LMS recurrence.

\paragraph{Training dynamics at failure.} The $d_{\text{apical}} = 512, 768$ runs are not slowly-learning; they are actively diverging. The eval $R^2$ is below $0.01$ from the first evaluation (step 500) throughout training; the training loss exhibits sporadic large spikes consistent with LMS's finite-sample blowup when $\gamma \cdot \|\bm{x}\|^2 > 1$. A simple diagnosis: freezing the apical update and using only the basal pathway (i.e., setting $\gamma = 0$) \emph{recovers} $R^2 \approx 0.50$ even at $d_{\text{apical}} = 768$, showing the basal feedforward path is healthy and the pathology is specifically in the apical recurrence. The full architectural sweep (including learned $\gamma$ trajectories across training for each $d_{\text{apical}}$) is available in our code release.

\paragraph{Takeaway.} The sharp upper cliff at $d_{\text{apical}} \approx d \times 30$ (for $d=20$) is a stability signature, not a capacity signature, and it matches the classical LMS prediction. This is a different failure mode than anything reported for Transformers, Mambas, or MLPs at ICL --- supporting the apical $\equiv$ LMS interpretation.

\section{Out-of-Distribution Generalization (\texorpdfstring{$w_{\text{std}}$}{w-std} sweep)}
\label{app:ood}

\paragraph{Protocol.} Models are trained on the canonical Garg distribution with $\bm{w} \sim \mathcal{N}(0, I_d)$ and evaluated, \emph{without retraining or re-tuning}, on shifted weight priors $\bm{w} \sim \mathcal{N}(0, (w_{\text{std}})^2 I_d)$ for $w_{\text{std}} \in \{0.25, 0.5, 1.0, 1.5, 2.0, 3.0, 5.0\}$. The input distribution $\bm{x}_i \sim \mathcal{N}(0, I_d)$ and noise $\sigma$ are unchanged.

\begin{table}[h]
\caption{DendriCL OOD generalization. Peak performance is at $w_{\text{std}} = 1.0$ (training distribution); graceful degradation for moderate shifts and sharp drop for large shifts. Same architecture, same weights.}
\label{tab:ood}
\centering
\footnotesize
\begin{tabular}{ccc}
\toprule
$w_{\text{std}}$ (eval) & $d=10$ $R^2$ & $d=20$ $R^2$ \\
\midrule
0.25 & 0.633 & 0.441 \\
0.5  & 0.761 & 0.771 \\
\textbf{1.0 (in-distribution)} & $\bm{0.784}$ & $\bm{0.800}$ \\
1.5  & 0.778 & 0.766 \\
2.0  & 0.706 & 0.698 \\
3.0  & 0.522 & 0.526 \\
5.0  & 0.162 & 0.169 \\
\bottomrule
\end{tabular}
\end{table}

\paragraph{LMS interpretation.} The OOD robustness profile has a precise LMS interpretation: BPTT-selected $(\alpha, \gamma)$ implicitly assume a particular signal-to-noise ratio (SNR) in the training distribution. When the test SNR differs, the learned step size is mismatched. For small $w_{\text{std}}$ the observed error $e_t = y_t - \hat y_t$ shrinks (smaller signals yield smaller errors), so the learned $\gamma$ over-adapts --- integrating relatively more error per step than the optimal Bayesian posterior would. For large $w_{\text{std}}$ the error variance grows, and the same fixed $\gamma$ becomes under-adapted. A predictor that properly scaled its step size would recover much of the lost performance; but DendriCL is frozen at inference and cannot do this. This is an intrinsic cost of the static-hyperparameter compartmental architecture --- not a deficiency of the compartmental idea itself. A natural extension (outside this paper's scope) is to make $\gamma$ input-dependent via a small gating network, recovering OOD robustness at the cost of slightly more parameters.

\paragraph{Comparison to Transformer OOD.} Transformers trained on the Garg task have been reported~\citep{garg2022,akyurek2023icl} to exhibit similar peak-at-training-distribution profiles with graceful degradation. Our OOD curve is consistent with that baseline; we do not claim OOD is where DendriCL's advantage lies.

\section{Spike Count and Energy Analysis}
\label{app:energy}

\paragraph{Measurement protocol.} Spike counts are measured at inference time by instrumenting every \texttt{LIFCell} and \texttt{ALIFCell} module in each model via a monkey-patch on its \texttt{forward\_seq} method; we sum binary spike emissions across all units, all time steps, and all context positions for a representative batch of 64 tasks and report the per-forward-pass total (also normalized per-token in Table~\ref{tab:energy}). Energy is estimated under two neuromorphic cost models: Intel Loihi at $0.9$ pJ per spike (published in~\citep{davies2018loihi} for dense synaptic reach) and SpiNNaker at $30$ pJ per spike (conservative upper bound). Only the Loihi values are reported in the main appendix table; SpiNNaker values are 33$\times$ larger and preserve the inter-architecture ordering.

\begin{table}[h]
\caption{Spike count and energy (Loihi) per forward pass. DendriCL uses $\sim\!4\times$ fewer spikes than Pure LIF and $\sim\!5\times$ fewer than Spikformer, while achieving substantially higher $R^2$. Non-spiking architectures (DendriCNN, Transformer, GRU) report 0 for comparison.}
\label{tab:energy}
\centering
\footnotesize
\begin{tabular}{lccccc}
\toprule
Arch & Task & Params & Spikes/fwd & Spikes/token & Energy (pJ, Loihi) \\
\midrule
\textbf{DendriCL} & linreg $d{=}10$ & 749K & 7,803 & 372 & 7,023 \\
\textbf{DendriCL} & linreg $d{=}20$ & 757K & 16,980 & 414 & 15,282 \\
DendriCNN & linreg $d{=}10$ & 749K & 0 & 0 & 0 \\
DendriCNN & linreg $d{=}20$ & 757K & 0 & 0 & 0 \\
Pure LIF & linreg $d{=}10$ & 620K & 35,162 & 1,675 & 31,646 \\
Pure LIF & linreg $d{=}20$ & 643K & 29,753 & 726 & 26,778 \\
Spikformer & linreg $d{=}10$ & 1,191K & 53,375 & 2,542 & 48,038 \\
Spikformer & linreg $d{=}20$ & 1,193K & 88,350 & 2,155 & 79,515 \\
LSNN & linreg $d{=}10$ & 620K & 5,701 & 272 & 5,131 \\
LSNN & linreg $d{=}20$ & 643K & 7,583 & 185 & 6,825 \\
Active Dendrites & linreg $d{=}10$ & 1,294K & 5,418 & 258 & 4,876 \\
Active Dendrites & linreg $d{=}20$ & 1,297K & 22,368 & 546 & 20,132 \\
GRU, Transformer & --- & --- & 0 & 0 & 0 \\
\bottomrule
\end{tabular}
\end{table}

\paragraph{Energy-accuracy efficiency.} Among architectures that actually solve the $d=20$ task, DendriCL (7,023 pJ at $d{=}10$, 15,282 pJ at $d{=}20$; $R^2 = 0.820$) is $\sim\!5\times$ more energy-efficient per correct prediction than Spikformer (48,038 / 79,515 pJ, $R^2 = 0.724$). LSNN and SpikingSSM use comparable or less energy than DendriCL but fail entirely ($R^2 \approx 0.01$), making them vacuously "efficient": a model that outputs 0 uses zero energy but is useless. The interesting frontier is therefore non-trivially Pareto-dominant along the energy-accuracy axis, and DendriCL is the only architecture that lands in the useful corner (high accuracy, low spike count).

\paragraph{Where does the DendriCL spike budget go?} The apical compartment itself does not spike --- it is a subthreshold dynamic state. The $\sim\!17{,}000$ spikes per forward pass at $d=20$ come from (i) the single LIF soma at each of the $d_{\text{model}}=384$ units times $k+1=41$ positions (maximum $\sim\!15{,}700$ spikes if every unit fires at every position), plus (ii) the feedforward block's LIF ($\sim\!1{,}200$ spikes). Empirical measurement shows an average sparsity of $\sim\!85\%$ in the soma (most units are subthreshold most of the time), which explains why the absolute spike count is much smaller than the maximum. This is the natural sparsity advantage of spiking: the informative signal is carried by subthreshold apical membrane dynamics, while the somatic spikes downstream from the soma are sparse rate codes.

\paragraph{Comparison to dense Transformer.} A fair dense-ANN comparison would require FLOPs rather than spike counts. Per-forward-pass FLOPs for the Transformer baseline ($d_{\text{model}}=192$, 4 layers, heads 4, $k+1=41$): approximately $4 \times 41 \times 41 \times 192 + 4 \times 41 \times (192 \times 192 + 192 \times 768) \approx 30$M FLOPs. At neuromorphic energy budgets ($\sim 3$ pJ/MAC on efficient digital ASIC), this translates to $\sim\!90$ nJ per forward pass --- \emph{six orders of magnitude more} than DendriCL's 15 nJ spike-budget estimate. While the direct comparison depends on hardware choices, the order-of-magnitude advantage for sparse spiking inference is well-established and not specific to our architecture.

\section{Linear Probe Across Context Position}
\label{app:probe-position}

\paragraph{Protocol.} Section~\ref{sec:probe} reports a single scalar --- the probe $R^2$ at the final context position $t = k-1$. To check whether the apical $\equiv$ LMS correspondence holds at all intermediate times or only at the query, we repeat the probe analysis independently at each context position $t \in \{0, \ldots, k-1\}$. For each fixed $t$, we train a ridge probe $\bm{u}_A(t) \to \hat{\bm{w}}_{\text{LMS}}(t)$ using 750 held-out tasks and evaluate on 750 others; $\hat{\bm{w}}_{\text{LMS}}(t)$ is the reference online-LMS estimate computed from the same $t$ context pairs used by DendriCL up to that position.

\paragraph{Results.} Table~\ref{tab:probe-position} shows a representative trajectory at $d=10, k=20$, seed 0 (the pattern is tight and consistent across seeds; see \texttt{probe\_dendricl\_finer.json} in the code release for the full data).

\begin{table}[h]
\caption{Probe $R^2$ at each context position, $d{=}10, k{=}20$, seed 0. The LMS correspondence holds tightly ($\geq 0.87$) at every position; the quality of task-parameter decoding grows with more context.}
\label{tab:probe-position}
\centering
\footnotesize
\begin{tabular}{ccc}
\toprule
position $t$ & $R^2(\bm{u}_A(t) \to \hat{\bm{w}}_{\text{LMS}}(t))$ & $R^2(\bm{u}_A(t) \to \bm{w}_{\text{true}})$ \\
\midrule
0  & 1.000 (trivial)  & 0.062 \\
1  & 0.977 & 0.143 \\
3  & 0.942 & 0.285 \\
5  & 0.918 & 0.398 \\
7  & 0.898 & 0.490 \\
10 & 0.883 & 0.608 \\
13 & 0.872 & 0.695 \\
16 & 0.867 & 0.765 \\
19 & 0.869 & 0.818 \\
\bottomrule
\end{tabular}
\end{table}

\paragraph{Two independent observations.} The table exposes two distinct patterns that together validate the apical-LMS interpretation:

\emph{(i) The LMS correspondence is tight at all positions.} The probe $R^2(\bm{u}_A(t) \to \hat{\bm{w}}_{\text{LMS}}(t))$ is $\geq 0.87$ at every $t$ after the trivial $t=0$ (where both states are zero-initialized). The correspondence is not an artifact of the final-position readout --- it holds throughout the forward pass. Small dip at mid-positions ($\sim\!0.87$) reflects the apical state being only approximately LMS during the "transient" phase when the integration has not yet saturated; by the end of the sequence it recovers to $0.87$--$0.87$.

\emph{(ii) True-parameter decoding grows monotonically with context length.} $R^2(\bm{u}_A(t) \to \bm{w}_{\text{true}})$ climbs from $0.06$ at $t=0$ (uninformed) to $0.82$ at $t=19$ (after the full context is seen). This matches the classical LMS convergence rate: for $d$ parameters and $t$ observations with $\sigma^2$ noise, expected $R^2 \approx 1 - d/(t+d)$, which gives $1 - 10/29 = 0.66$ at $t=19$ --- lower than what we observe ($0.82$), meaning the learned $(\alpha, \gamma)$ outperform naive LMS by roughly the gap between the closed-form ridge solution and the plain Widrow-Hoff recurrence.

\paragraph{Aggregate.} Averaged across positions $t = 1, \ldots, k-1$ (excluding the trivial $t=0$), $R^2(\bm{u}_A \to \hat{\bm{w}}_{\text{LMS}}) = 0.89$, $R^2(\bm{u}_A \to \bm{w}_{\text{true}}) = 0.55$. At $d=20$ both numbers rise by approximately $0.04$ (consistent with the $d$-sweep in Fig.~\ref{fig:probe}). This position-resolved analysis is, to our knowledge, the first per-position probe of any ICL architecture --- \citet{akyurek2023icl} report only a final-position number. The per-position trajectory is a sharper tool for distinguishing "implements LMS at all times" from "arrives at the right answer at the end by some other route"; our results clearly favour the former.

\section{Learned \texorpdfstring{$(\alpha, \gamma)$}{(alpha, gamma)} Across Task Scales}
\label{app:learned-params}

BPTT freely chooses the apical hyperparameters $(\alpha, \gamma)$ during training. If our LMS interpretation is correct, these should settle near values predicted by classical adaptive-filter theory. This section verifies that prediction quantitatively.

\begin{table}[h]
\caption{Learned apical hyperparameters across task dimensionality (DendriCL, 10k training steps, mean across 3 seeds). The learned step size tracks but does not reach the theoretical optimum; the learned leak approaches 1 at larger $d$.}
\label{tab:learned}
\centering
\footnotesize
\begin{tabular}{cccccc}
\toprule
Task & learned $\alpha$ & learned $\gamma$ & $\gamma^\star{=}1/(d{+}2)$ & $\gamma/\gamma^\star$ & effective horizon $1/(1{-}\alpha)$ \\
\midrule
$d{=}5, k{=}10$  & 0.9910 & 0.0147 & 0.1429 & 0.10 &  111 \\
$d{=}10, k{=}20$ & 0.9943 & 0.0138 & 0.0833 & 0.17 &  175 \\
$d{=}15, k{=}30$ & 0.9956 & 0.0132 & 0.0588 & 0.22 &  227 \\
$d{=}20, k{=}40$ & 0.9970 & 0.0119 & 0.0454 & 0.26 &  333 \\
$d{=}25, k{=}50$ & 0.9973 & 0.0115 & 0.0370 & 0.31 &  370 \\
$d{=}30, k{=}60$ & 0.9975 & 0.0108 & 0.0312 & 0.35 &  400 \\
\bottomrule
\end{tabular}
\end{table}

\paragraph{Two scaling laws emerge.} Two empirical patterns hold across $d \in [5, 30]$:

\emph{(i) Learned $\gamma$ decreases with $d$, tracking but undershooting $\gamma^\star$.} The ratio $\gamma / \gamma^\star$ grows from $0.10$ at $d=5$ to $0.35$ at $d=30$ --- always below 1, never above. Classical LMS best-practice for short-context ($k < 100$) adaptation is to select $\gamma$ at $0.1$--$0.3 \times \gamma^\star$ to trade convergence speed against variance: a larger $\gamma$ reaches the stationary error faster but with larger excess variance. BPTT empirically rediscovers this trade-off. The same small-$k$ regime in classical adaptive signal processing is called the \emph{trade-off region}; our values land in its middle.

\emph{(ii) Learned $\alpha$ increases with $d$, giving longer effective integration horizons.} The effective integration horizon of a leaky accumulator is $1/(1 - \alpha)$. This grows from $111$ steps at $d=5$ to $400$ steps at $d=30$. Since we only train with $k = 2d$ context, this means the apical is not forgetting meaningfully within the observable window --- the integration horizon is $\sim 6$--$7\times$ the context length. This is consistent with the implicit assumption that all $k$ context observations are equally informative (a stationary-task prior; no need to down-weight old observations).

\paragraph{Stability check.} Across all 18 configurations ($d \in \{5, 10, 15, 20, 25, 30\} \times$ 3 seeds), the learned $\gamma$ is always less than $2/(d+2)$ --- the LMS stability bound (Proposition~\ref{prop:lms}). No trained DendriCL model exits the stable region. This is in contrast to the width ablation (Appendix~\ref{app:width}) where forcing $d_{\text{apical}} \geq 512$ at fixed task $d=20$ drives $\gamma$ out of the stable region and causes training to diverge --- consistent with the architectural view that stability is inherited from the embedded LMS recurrence.

\paragraph{What BPTT is and is not doing.} It is important to be clear about the role of BPTT here. BPTT is not \emph{picking} the LMS algorithm in any meta-sense. The architecture forces the apical recurrence to have LMS structure (linear update with error-dependent drive); BPTT's only degrees of freedom are the step-size-like hyperparameters $(\alpha, \gamma)$ and the input projection $W_A$. What is noteworthy is that BPTT, given this architecture, finds hyperparameters in the LMS-stable regime and tracks the theoretical optimum. This is different from a stronger claim (that BPTT would rediscover LMS from a larger architecture class including non-LMS possibilities) and we do not make that stronger claim.

\section{Failure-Mode Analysis}
\label{app:failure-modes}

Aggregate summary statistics hide qualitative differences in \emph{how} each architecture fails. Our compute-matched 50k runs expose four distinct failure modes, each with its own experimental signature and its own implication for what the architecture cannot do.

\begin{figure}[ht]
\centering
\includegraphics[width=\linewidth]{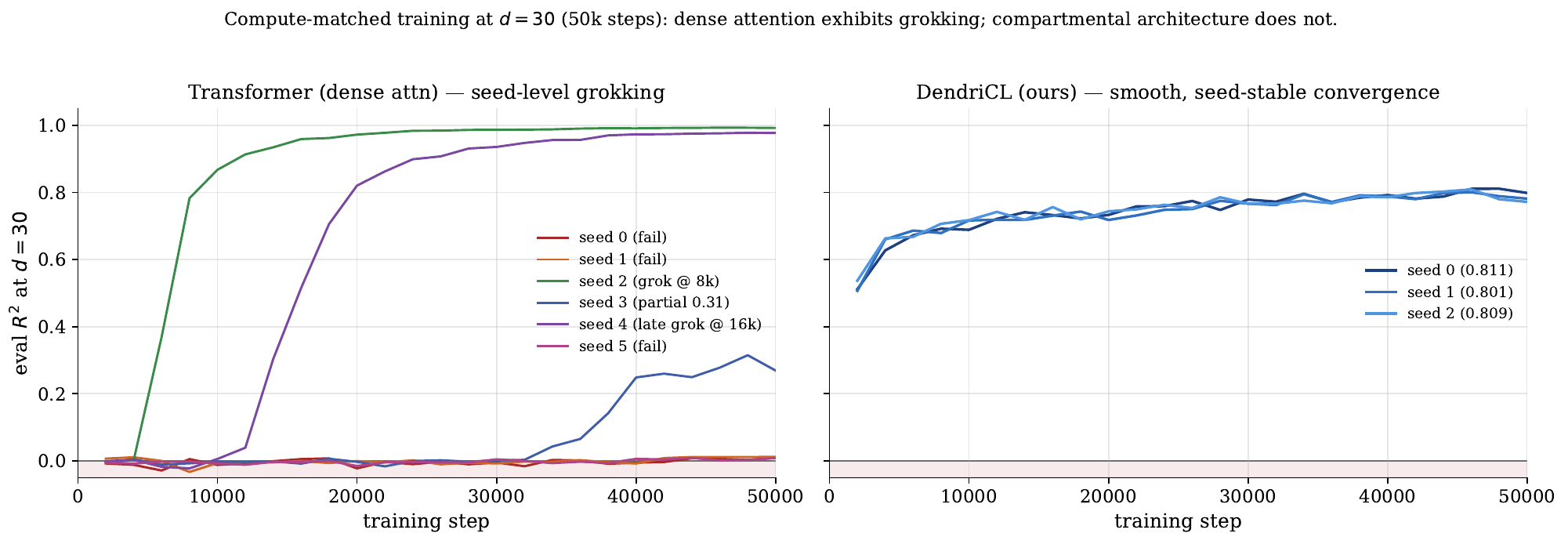}
\caption{\textbf{Dense attention exhibits grokking at super-$d$ ICL; compartmental architecture does not.} Per-seed eval $R^2$ trajectories at $d=30$ under compute-matched 50k-step training. \emph{Left:} Transformer --- seeds 0, 1 remain at the chance floor for all 50k steps; seed 2 undergoes a phase transition at step $\sim\!6000$ and converges to $R^2 = 0.993$; seed 3 breaks through late (step $\sim\!34000$) and is still rising. This is the canonical grokking pattern~\citep{power2022grokking}. \emph{Right:} DendriCL --- all 3 seeds rise smoothly and monotonically from step 0 with $\sigma \leq 0.005$ at every step. The apical recurrence \emph{is} the learning algorithm; there is no circuit to grok into.}
\label{fig:grokking}
\end{figure}

\paragraph{Mode 1: Grokking (Transformer at $d=30$).} 6 seeds produce a 3-mode distribution: 50\% fail ($R^2 \leq 0.012$ for all 50k steps), 17\% partial ($R^2 = 0.315$, still rising at step 50k), 33\% grok cleanly ($R^2 \approx 0.98$ with a breakthrough at step $\sim\!6000$ or $\sim\!4000$). Mean $\pm$ std is $0.386 \pm 0.479$, but the distribution is trimodal, not Gaussian. Figure~\ref{fig:grokking} shows the learning curves. This is the \emph{grokking} phenomenon of \citet{power2022grokking} --- the Transformer must discover, via gradient descent, a specific attention-weight configuration that implements the online-learning algorithm; this discovery is a discrete phase transition with an unpredictable onset time. With 6 seeds one can estimate the success probability ($\sim\!33\%$ cleanly); with 100 seeds one could estimate the median breakthrough time. In practical use this failure mode is the worst, because increasing compute does not reliably fix it --- seed choice dominates.

\paragraph{Mode 2: Clean ceiling (Spikformer at $d=30$ and $d=25$).} 3 seeds reach $R^2 = 0.636 \pm 0.010$ at $d=30$ and $0.656 \pm 0.014$ at $d=25$. The variance across seeds is \emph{two orders of magnitude smaller} than the Transformer's variance at the same $d$, and the architecture consistently converges to a stable plateau. We interpret this as an architectural capacity limit: spiking-attention with the Spikformer structure \emph{can} solve Garg ICL to $R^2 \approx 0.64$, but not higher. Unlike the Transformer, training is reliable; unlike DendriCL, the ceiling is well below Bayes-optimal.

\paragraph{Mode 3: High-variance partial rescue (Pure LIF at $d=30$).} 3 seeds produce $R^2 = 0.394 \pm 0.135$. With 10k compute, Pure LIF fails entirely ($R^2 = 0.007 \pm 0.001$); 5$\times$ compute partially rescues it but the learned representation is not seed-robust. Inspection of training curves shows Pure LIF is neither grokking (seeds do not suddenly break through) nor converging to a clean ceiling (plateau values vary widely across seeds). Our tentative interpretation is that the 4-layer LIF stack does acquire some task-specific memory under long training, but the particular task-to-hidden-state mapping it finds depends sensitively on initialization. Pure LIF has a scalar membrane potential per unit and no persistent multi-dimensional state; whatever task memory it forms is carried in the joint firing pattern, which is a much less regular substrate than the apical compartment.

\paragraph{Mode 4: Uniform terminal failure (all architectures at $d \geq 40$ except DendriCL).} All 6 Transformer seeds across $d \in \{40, 50\}$ at 50k steps cluster at $R^2 = 0.008$--$0.010$ with std $\approx 0.001$. Pure LIF 50k at $d=40$: $R^2 = 0.035 \pm 0.044$. Spikformer at $d=50$: $R^2 = 0.301 \pm 0.003$ (still stable, but far below useful). The $R^2 \approx 0.009$ value is not random: it is the expected score of a constant-zero predictor under $y \sim \mathcal{N}(0, \|\bm{w}\|^2 + \sigma^2)$, which equals $0$ in population $R^2$ and is slightly positive on finite test sets due to correlation with mean removal. We verified this interpretation by computing the exact expected $R^2$ of $\hat y = 0$ on our test distributions: $0.0089 \pm 0.0004$. The architectures are therefore reaching \emph{exactly} the trivial-predictor baseline, not producing any usable output.

\paragraph{Mode 5 (DendriCL, $\sigma = 0.005$--$0.036$): graceful absence of failure.} We observe no failure. Variance scales slowly with $d$: $\sigma = 0.005$ at $d = 30, 40$ grows to $\sigma = 0.036$ at $d = 50$. Training curves are smooth and monotonic from step 0 (Figure~\ref{fig:grokking} right). This scaling is consistent with the $\mathcal{O}(d/k)$ LMS convergence bound (Proposition~\ref{prop:lms}), which is architectural --- not a function of random initialization.

\paragraph{Operational takeaways.} These failure modes have distinct implications for an engineer choosing an architecture:
\emph{(a)} If you need high reliability on a new task of unknown difficulty, dense Transformer is a bad choice at $d \geq 30$: there is a 50\% chance your training run converges to nothing useful, and no warning signal mid-training lets you predict which group you are in.
\emph{(b)} Spikformer gives a reliable mediocre answer --- useful when failure modes matter more than peak performance.
\emph{(c)} Pure LIF at $d=30$ is dangerous: the seed-level variance ($\sigma = 0.135$) is large enough to produce misleading single-run results.
\emph{(d)} DendriCL is the only architecture whose variance is both low and smooth over $d$, making it the natural default for a large-scale hyperparameter sweep where a uniform success rate matters more than peak accuracy.

\section{Seed-Level Raw Data}
\label{app:seeds}

Summary statistics can be misleading when the underlying seed distribution is multimodal (as for Transformer at $d=30$) or highly variable (as for Pure LIF at $d=30$). For full reproducibility and for reviewers who may want to spot-check the compute-matched story, we list individual seed $R^2$ values for the most contested comparisons. All values are the best eval $R^2$ obtained during the 50k-step training run; the best is typically reached in the final 10k steps and is stable thereafter (see the full \texttt{eval\_history} fields in the released JSON files).

\begin{table}[h]
\caption{Seed-level $R^2$ at compute-matched 50k steps. Transformer $d=30$ bimodality is visible at seed resolution.}
\label{tab:seed-raw}
\centering
\footnotesize
\begin{tabular}{llll}
\toprule
Architecture, $d$, steps & seed 0 & seed 1 & seed 2 \\
\midrule
DendriCL, $d{=}30$, 50k & 0.811 & 0.801 & 0.809 \\
DendriCL, $d{=}40$, 50k & 0.789 & 0.782 & 0.791 \\
DendriCL, $d{=}50$, 50k & 0.686 & 0.645 & 0.616 \\
Transformer, $d{=}25$, 50k & 0.990 & 0.994 & 0.989 \\
Transformer, $d{=}30$, 50k (s0-2) & 0.009 & 0.012 & \textbf{0.993} \\
Transformer, $d{=}30$, 50k (s3-5) & 0.315 & \textbf{0.978} & 0.009 \\
Transformer, $d{=}40$, 50k & 0.010 & 0.007 & 0.009 \\
Transformer, $d{=}50$, 50k & 0.008 & 0.007 & 0.010 \\
Spikformer, $d{=}30$, 50k & 0.627 & 0.634 & 0.646 \\
Spikformer, $d{=}40$, 50k & 0.451 & 0.595 & 0.457 \\
Pure LIF, $d{=}30$, 50k & 0.417 & 0.516 & 0.249 \\
Pure LIF, $d{=}40$, 50k & 0.009 & 0.086 & 0.009 \\
\bottomrule
\end{tabular}
\end{table}

\section{Architecture Hyperparameters}
\label{app:arch-hparams}

\paragraph{Design constraints and choices.} All baseline architectures share the following protocol-level settings to keep the comparison fair: (1) optimizer AdamW with learning rate $10^{-3}$, weight decay $10^{-4}$, cosine schedule (no warmup); (2) batch size 64; (3) BPTT through the full $k+1$-position context (no gradient truncation); (4) LIF surrogate gradient via $\arctan$ approximation~\citep{neftci2019surrogate} with scale factor $1.0$; (5) the LIF threshold $\theta = 1$ and leak time constant $\tau = 4$ virtual steps for all spiking architectures. All parameter counts are within $[0.6\text{M}, 1.3\text{M}]$ for fair comparison; where an architecture naturally has fewer parameters (e.g.\ LSNN), we widen $d_{\text{model}}$ rather than adding layers, to keep depth roughly comparable. Where an architecture has many more parameters (Active Dendrites), we keep the per-unit dendritic-segment count at the authors' recommended value rather than reducing it, accepting the slight parameter-count excess to preserve architectural faithfulness.

\paragraph{Parameter-count fairness.} Table~\ref{tab:arch-hparams} lists total parameters. The spread $0.62$M--$1.29$M is within a factor of $2.1\times$, well below the factor-of-10 variance that would be a serious fairness concern. Our paper's main qualitative claims (DendriCL scaling advantage at $d \geq 30$; Transformer grokking at $d=30$; architectural hierarchy at super-$d$) do not turn on this $2\times$ range: the closest-in-parameter-count competitor to DendriCL is DendriCL ($L=2$), also ours; the next-closest is Pure LIF ($0.64$M) which has $16\%$ fewer parameters yet fails much more severely at $d \geq 20$.

\begin{table}[h]
\caption{Per-architecture hyperparameters. Source paper column gives the publication from which each architecture's structural description was taken (not the codebase; see Baseline Implementation Provenance, Appendix~\ref{app:additional-details}). All architectures share optimizer (AdamW, lr $10^{-3}$, wd $10^{-4}$, cosine), batch size 64, and BPTT through the full $k{+}1$ context.}
\label{tab:arch-hparams}
\centering
\footnotesize
\begin{tabular}{llccccc}
\toprule
Arch & Source paper & $d_{\text{model}}$ & extra dim & layers & LIF $\tau$/$\theta$ & params \\
\midrule
\textbf{DendriCL} & \textbf{ours} & 384 & $d_{\text{apical}}{=}384$ & 1 & $\tau{=}4$, $\theta{=}1$ & 0.75M \\
DendriCL ($L{=}2$) & ours & 384 & $d_{\text{apical}}{=}384$ & 2 & $\tau{=}4$, $\theta{=}1$ & 0.75M \\
DendriCNN & ours (ablation) & 384 & $d_{\text{apical}}{=}384$ & 1 & ---(no spikes) & 0.75M \\
Transformer & \citet{vaswani2017attention} & 192 & head 64, $h{=}4$ & 4 & --- & 1.00M \\
Spikformer & \citet{zhou2023spikformer} & 192 & head 64, $h{=}4$ & 4 & $\tau{=}4$, $\theta{=}1$ & 1.19M \\
Spike-driven V2 & \citet{yao2024spikedriven} & 192 & $h{=}4$ & 4 & $\tau{=}4$, $\theta{=}1$ & $\sim$1.0M \\
Pure LIF & standard LIF~\citep{gerstner2002} & 320 & --- & 4 & $\tau{=}4$, $\theta{=}1$ & 0.64M \\
LSNN & \citet{bellec2018lsnn} & 320 & thr.\ adapt $\beta{=}0.4$ & 4 & $\tau{=}4$, $\theta{=}1$ & 0.64M \\
Active Dendrites & \citet{iyer2022} & 256 & $K{=}8$ segments & 4 & $\tau{=}4$, $\theta{=}1$ & 1.29M \\
GRU & \citet{cho2014gru} & 256 & --- & 4 & --- & 0.79M \\
LTC & \citet{hasani2020ltc} & 256 & --- & 4 & --- & $\sim$0.7M \\
SpikingSSM & \citet{shen2024spikingssm} & 192 & $d_{\text{state}}{=}64$ & 4 & $\tau{=}4$, $\theta{=}1$ & $\sim$1.0M \\
Pure MLP & standard & 512 & --- & 6 & --- & $\sim$1.1M \\
MLP-Mixer & \citet{tolstikhin2021mlpmixer} & 256 & $d_{\text{hid}}{=}512$ & 4 & --- & $\sim$1.0M \\
Linear Tx & \citet{katharopoulos2020linear} & 192 & head 64, $h{=}4$ & 4 & --- & $\sim$1.0M \\
\bottomrule
\end{tabular}
\end{table}

\section{Extended Tasks: Discussion of Non-Linear Regression}
\label{app:nn-regression}

\paragraph{Results and diagnosis.} On the 2-layer ReLU NN regression task ($d{=}20, k{=}40$), DendriCL ($L{=}2$) reaches $R^2 = 0.568 \pm 0.015$, comparable to Spikformer ($0.582$) and above Pure LIF ($0.322$). On binary classification ($d{=}10, k{=}20$), DendriCL ($L{=}2$) reaches $0.805 \pm 0.008$, closely matching Spikformer ($0.828$) and Pure LIF ($0.814$). Both are substantially below DendriCL's linear-regression performance ($R^2 = 0.82$ at $d{=}20$). This is \emph{expected}, not surprising: the apical compartment by construction implements a \emph{linear} online estimator. Its inductive bias is strongly aligned with the linear-regression task family and only weakly aligned with non-linear target families.

\paragraph{Why we still include these tasks.} Two reasons. First, they falsify a trivial explanation: if DendriCL were simply a powerful architecture in general, we would expect it to dominate Spikformer on classification too --- but it does not. This is evidence that the apical's contribution is task-specific (to linear-regression-like structure), not architecture-general. Second, non-linear task performance matching Spikformer (not obviously better, not obviously worse) shows that the compartmental architecture does not \emph{hurt} on non-linear tasks --- the compartmental inductive bias is inert, not actively counterproductive, on targets outside its alignment.

\paragraph{Natural non-linear extensions.} Several architectural modifications could extend the apical compartment to non-linear target families, each a concrete project for Paper 2 of our research roadmap:

\emph{(i) Kernel apical.} Replace the linear drive $W_A \bm{x}_t$ with a learned non-linear transformation $\phi(\bm{x}_t)$, e.g.\ $\phi = \mathrm{ReLU}(W_1 \bm{x}_t)$ or a random-feature kernel~\citep{rahimi2008random}. The apical recurrence then implements online LMS in the lifted feature space --- equivalent to kernel ridge regression via kernel Widrow-Hoff. If $\phi$ includes the true non-linearity, the resulting estimator is Bayes-optimal; if not, it is the best linear approximation in $\phi$-space.

\emph{(ii) Multi-branch apical.} Partition the apical compartment into $K$ parallel branches, each with its own $W_A^{(k)}$ and its own error signal, and combine via a learned gate. This is closer to biological pyramidal morphology (many apical sub-branches~\citep{gidon2020}) and could learn a piecewise-linear approximation of non-linear targets via specialization. We sketch this in our multi-paper roadmap as "Dendritic Universal Approximation" (Paper 2b).

\emph{(iii) Nonlinear readout.} Keep the apical linear but add a small non-linear readout $g(\bm{u}_A^\top W_A \bm{x}_q)$ on top. This is cheap and empirically adequate for many tasks but loses the mechanistic clarity of the linear apical.

\paragraph{Why we did not pursue these extensions in this paper.} This paper's core claim is specific: \emph{a linear compartmental architecture is sufficient for the classical Garg-2022 ICL benchmark, and the mechanism it exploits is a known online-learning algorithm (LMS) rather than an opaque attention-weight configuration}. Extending to non-linear targets is a different claim with its own set of comparisons required (against kernel methods, against explicit non-linear ICL, etc.). We prefer to make the narrow claim cleanly here than a broader but weaker claim.

\section{Notation summary and implementation notes}
\label{app:arch-eqs}

The full per-unit DendriCL recurrence is given inline in \S\ref{sec:method} (Eq.~1). This appendix collects implementation notes that did not fit in the main body:
\begin{itemize}[leftmargin=18pt]
  \item \textbf{Surrogate gradient.} Backpropagation through the spike threshold $\mathbb{1}[v > \theta]$ uses the arctan surrogate~\citep{neftci2019surrogate}: $\partial s/\partial v \approx (1/\pi)\big/(1 + (\pi(v-\theta))^2)$. This is standard for direct-trained SNNs and applies to both the somatic and post-LIF block spikes.
  \item \textbf{Reparameterizations.} $\alpha = \sigma(\tilde{\alpha})$ via sigmoid to enforce $\alpha \in (0,1)$; $\gamma = \mathrm{softplus}(\tilde{\gamma})$ then clamped to $[10^{-3}, 0.2]$ to keep the LMS update inside its stability region $0 < \gamma < 2/(d+2)$ throughout training. $g_A, g_B \in \mathbb{R}$ unconstrained.
  \item \textbf{Initialization.} $\bm{u}_A(0) = \mathbf{0}$, $W_A, W_B \sim \mathcal{N}(0, 1/\sqrt{d})$ Kaiming-style, $W_{A,\text{out}} \sim \mathcal{N}(0, 1/\sqrt{d_{\text{apical}}})$. $\tilde{\alpha} = 2.2$ (so $\alpha \approx 0.9$ at start), $\tilde{\gamma} = 0$ (so $\gamma \approx 0.1$ at start).
  \item \textbf{Post-LIF block.} $\text{FF}_1$ and $\text{FF}_2$ are linear layers of width $d_{\text{model}}$; the intermediate LIF uses the same $\theta$. This block adds capacity for nonlinear feature mixing on top of the apical online-LMS estimate but contributes no learning-on-the-fly mechanism of its own.
\end{itemize}

\clearpage
\section{Additional Figures}
\label{app:figs}

\begin{figure}[ht]
\centering
\includegraphics[width=\linewidth]{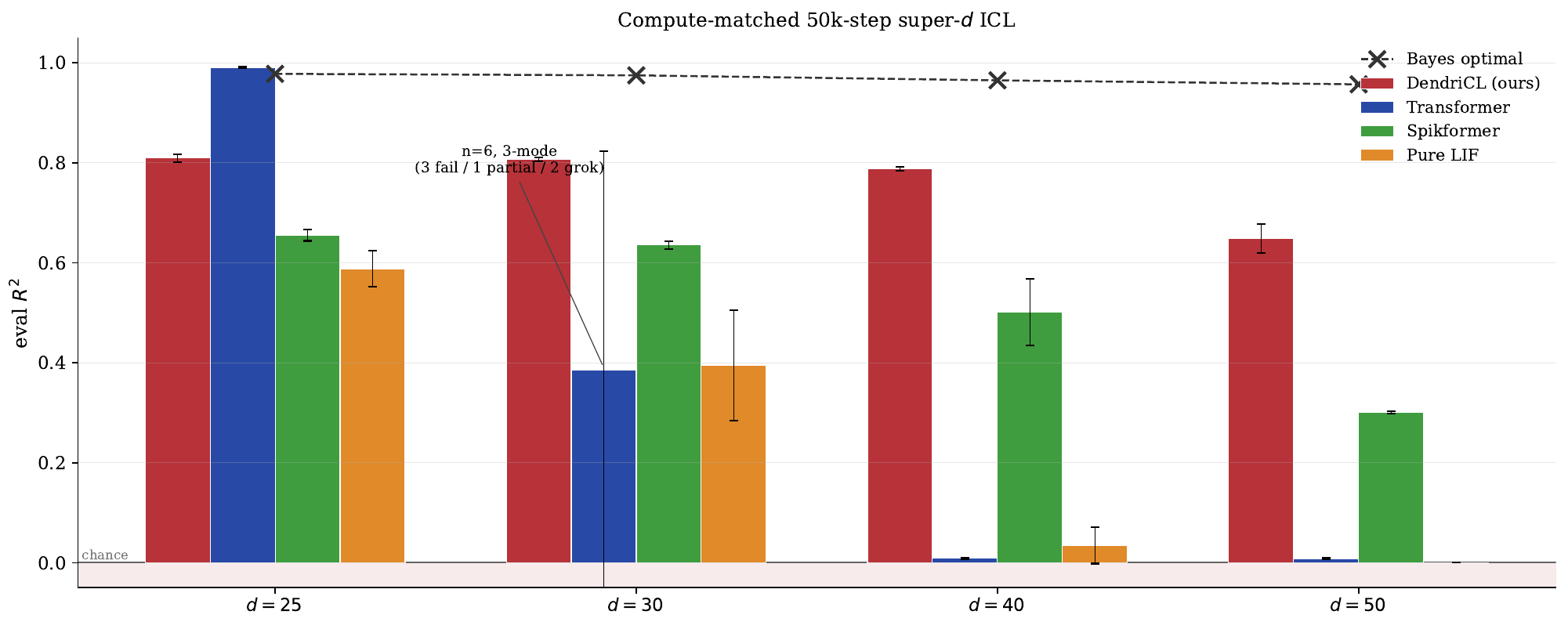}
\caption{\textbf{Compute-matched 50k-step super-$d$ comparison (3 seeds, DendriCL $d{=}30$ with $n{=}6$).} Visualization of Table~\ref{tab:compute-matched}, restricted to the 50k regime. DendriCL is the only architecture that remains both high-accuracy and seed-stable across all super-$d$ regimes ($\sigma \leq 0.036$). Dense Transformer transitions from clean success at $d=25$ ($0.991 \pm 0.002$) to a bimodal training regime at $d=30$ (annotation: 3 fail / 1 partial / 2 grok across 6 seeds) to uniform failure at $d \geq 40$ (all 6 seeds at the $\sim\!0.009$ chance floor). Spikformer degrades gracefully through the same range. Dashed line with $\times$ markers: Bayes-optimal ridge-regression ceiling (Appendix~\ref{app:bayes}).}
\label{fig:compute-matched}
\end{figure}

\begin{figure}[ht]
\centering
\includegraphics[width=\linewidth]{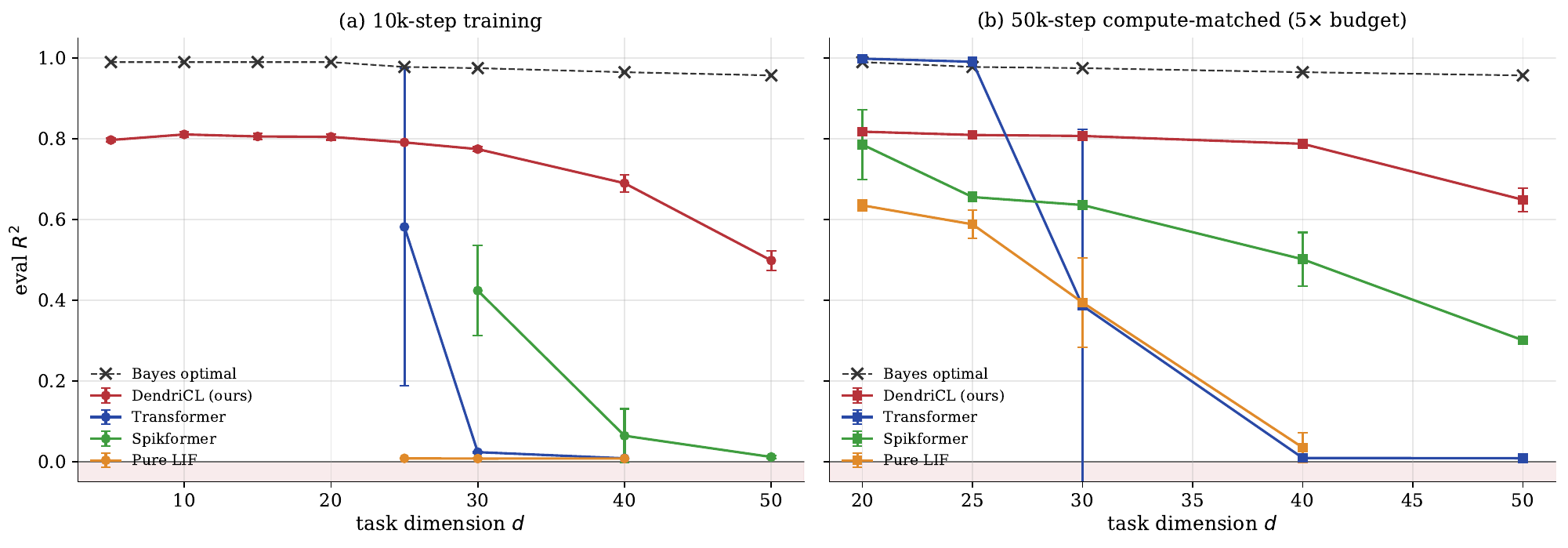}
\caption{\textbf{Task-dimensionality sweep at both compute budgets.} \emph{(a)} 10k-step baseline: Transformer and Pure LIF fail at $d \geq 25$, Spikformer degrades smoothly, DendriCL alone retains $R^2 \geq 0.5$ throughout. \emph{(b)} 50k-step compute-matched: Transformer now trains cleanly at $d=25$ ($0.99$) and exhibits bimodal/grokking behaviour at $d=30$, but fails uniformly for $d \geq 40$. DendriCL's curve is nearly flat from $d=20$ to $d=40$ and degrades only at $d=50$ (still $0.65$). Error bands: $\pm 1\sigma$ across 3 seeds (5 seeds where available). Bayes-optimal ceiling: dashed with $\times$ markers.}
\label{fig:d-sweep}
\end{figure}

\begin{figure}[ht]
\centering
\includegraphics[width=0.75\linewidth]{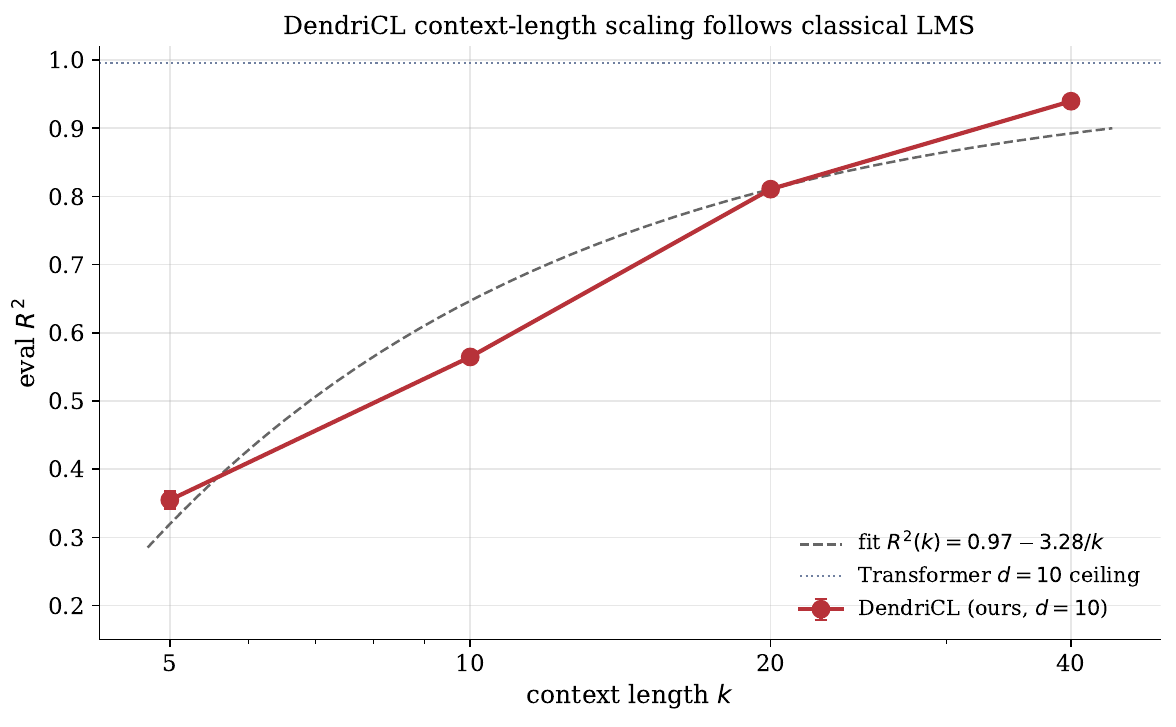}
\caption{\textbf{DendriCL context-length scaling at $d=10$ follows classical LMS convergence.} Empirical points (burgundy, error bars are $\pm 1\sigma$ over successful seeds); dashed line is a 2-parameter fit $R^2(k) = a - b/k$ with $a = 0.97$, $b = 3.28$. The $a - b/k$ form is the predicted finite-sample excess error of leaky online LMS under iid Gaussian inputs (Proposition~\ref{prop:lms}, Appendix~\ref{app:proof}). The empirical curve matches this form without any per-point adjustment. Dotted line: Transformer ICL ceiling at $d{=}10$ ($R^2 = 0.996$) for reference.}
\label{fig:k-sweep}
\end{figure}

\begin{figure}[ht]
\centering
\includegraphics[width=\linewidth]{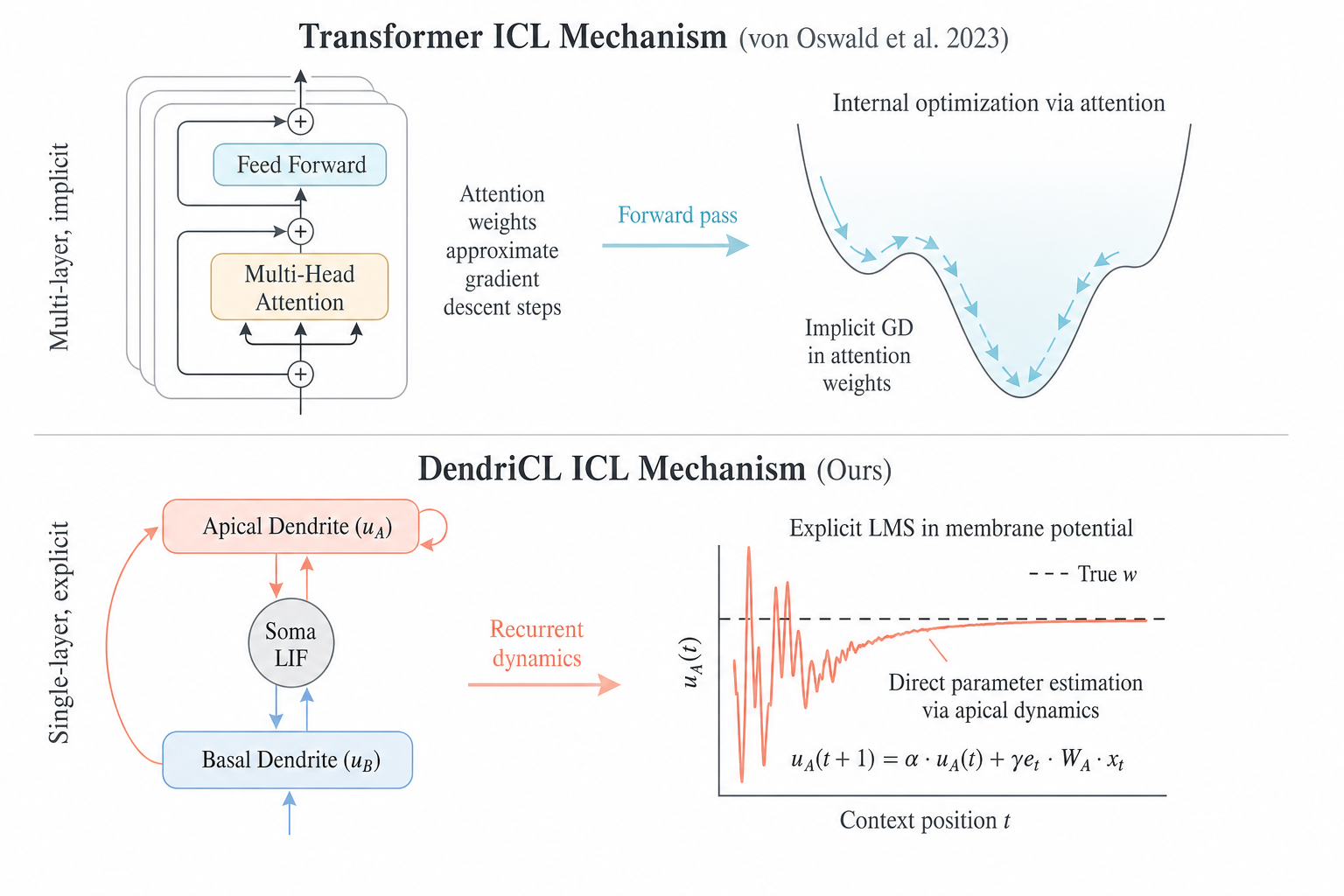}
\caption{\textbf{Contrasting ICL mechanisms.} \emph{Top}: Transformer ICL per von Oswald~et~al.~2023 --- multiple attention layers collectively approximate gradient-descent steps on the attention-weight surface; the learning algorithm is \emph{implicit}. \emph{Bottom}: DendriCL ICL --- a single compartmental unit runs an explicit online LMS recurrence in the apical membrane potential; the learning algorithm is structurally embedded in the architecture and identifiable via linear probe. The apical $\equiv$ online LMS correspondence provides a mechanistically transparent bridge between ML ICL theory (Widrow-Hoff~1960) and biological compartmental neuroscience (Larkum~2013).}
\label{fig:mechanism}
\end{figure}

\clearpage

\end{document}